\documentclass[runningheads]{llncs}

\usepackage{eccv}

\usepackage{eccvabbrv}

\usepackage{graphicx}
\usepackage{booktabs}

\usepackage[accsupp]{axessibility}  

\usepackage{graphicx}
\usepackage[most]{tcolorbox}
\usepackage{listings}
\usepackage{float}

\lstdefinestyle{systempromptstyle}{
    basicstyle=\ttfamily\scriptsize,
    breaklines=true,
    breakatwhitespace=false,
    columns=fullflexible,
    keepspaces=true,
    showstringspaces=false
}

\newtcolorbox{systempromptbox}{
    colback=white,
    colframe=black,
    arc=2mm,
    boxrule=0.6pt,
    left=2mm,
    right=2mm,
    top=2mm,
    bottom=2mm,
    width=\linewidth
}

\usepackage{subcaption}

\lstdefinestyle{promptstyle}{
    basicstyle=\ttfamily\tiny,
    breaklines=true,
    breakatwhitespace=false,
    columns=fullflexible,
    keepspaces=true,
    showstringspaces=false,
    aboveskip=0pt,
    belowskip=0pt
}

\newtcblisting{promptbox}{
    listing only,
    listing options={style=promptstyle},
    colback=white,
    colframe=black!25,
    boxrule=0.4pt,
    arc=1mm,
    left=1mm,
    right=1mm,
    top=1mm,
    bottom=1mm,
    boxsep=1mm,
    width=\linewidth,
    before skip=0.35em,
    after skip=0pt
}

\usepackage{subcaption}
\usepackage{placeins}

\usepackage{hyperref}

\usepackage{orcidlink}

\begin{document}

\title{AmalthAI: An Open-Source Computer Vision Platform for Cultural Heritage} 

\author{Christos Chatzisavvas\inst{1,2}\orcidlink{0009-0005-6274-8051} \and
Stelios Alvanos\inst{1,2}\orcidlink{0009-0002-0251-4630} \and
Efstratios Politis \inst{1,2}\orcidlink{0009-0000-8410-3477} \and
Panagiotis Rigas \inst{3,4}\orcidlink{0009-0006-5395-6112} \and
Thomas Pappas \inst{2}\orcidlink{0009-0008-5582-2024} \and
Ioannis Giannoukos \inst{2}\orcidlink{0009-0007-4965-1292} \and
Nikolaos Mitianoudis \inst{1, 2}\orcidlink{0000-0003-0898-6102} \and
Agata Ulanowska \inst{5}\orcidlink{0000-0002-4946-2711} \and
Katarzyna Żebrowska \inst{5}\orcidlink{0000-0002-9115-4165} \and
Nazarij Buławka \inst{5}\orcidlink{0000-0003-2338-4687} \and
Christina Margariti \inst{2}\orcidlink{0000-0002-3104-5397} \and
George Pavlidis \inst{2}\orcidlink{0000-0002-9909-1584} \and
Chairi Kiourt \inst{2}\orcidlink{0000-0001-8501-8899} \and
Anestis Koutsoudis \inst{2}\orcidlink{0000-0001-9862-7335} \and
Vassilis Katsouros \inst{2}\orcidlink{0000-0002-4185-2344} \and
George Ioannakis \inst{2}\orcidlink{0000-0001-6230-7030}} 

\authorrunning{C.~Chatzisavvas \etal}

\institute{
    Department of Electrical and Computer Engineering, Democritus University of Thrace, Greece \and
    Institute for Language and Speech Processing, Athena Research Center, Greece \and
    Department of Informatics \& Telecommunications, National and Kapodistrian University of Athens, Greece \and
    Archimedes, Athena Research Center, Greece \and Faculty of Archaeology, University of Warsaw, Poland
}

\maketitle

\begin{abstract}
    Computer vision (CV) and machine learning (ML) offer new tools for cultural heritage (CH) artifact analysis, but the CV/ML pipeline remains largely inaccessible to CH domain experts, who lack the background to configure, train, or assess models. We present \emph{AmalthAI}, an open-source CV platform that bridges this gap, enabling non-ML CH experts to independently produce and validate archaeologically meaningful findings. The interface covers dataset management, training, and inference for classification, segmentation, and object detection, with Kubeflow and Katib handling scalable training and hyperparameter search. Grad-CAM localizes the image region behind a prediction, and a vision-language model (VLM) adds a text description of it for expert review. Since archaeological data is often state-owned or rights-encumbered and cannot leave institutional custody, AmalthAI's self-hostable deployment ensures sensitive data is kept within premises. We test the platform on an archaeological use case built on a custom dataset of clay textile imprints, where CH experts trained and validated segmentation, and classification models for hypothesis testing. We provide the implementation code at \url{https://github.com/TEXTaiLES/AmalthAI}.

\end{abstract}
\section{Introduction}

Cultural heritage research increasingly depends on automated image analysis, from architectural classification~\cite{ottoni2025deep} to textile pattern recognition~\cite{rei2024multimodal}, a shift enabled by advances in deep learning and mature software libraries~\cite{paszke2019pytorch,abadi2016tensorflow}. The same holds for semantic segmentation of cultural heritage imagery~\cite{ragusa2021semantic}, a task CH researchers can define on archaeological grounds. Building computer vision applications, however, still requires expertise in dataset preparation, model configuration, parameter tuning, and software engineering, a barrier for CH researchers whose primary expertise lies outside computer science. 

CH research needs both a no-code interface and infrastructure controlled by the lab itself, yet existing platforms split the two. The ones open enough to self-host tend to target users already fluent in machine learning pipelines. The ones built for no-code accessibility tend to keep training and dataset management on vendor cloud infrastructure by default, with self-hosting locked behind enterprise contracts most CH research groups cannot afford or negotiate. Neither kind of platform closes the loop back to the domain expert. Evaluation often stops at accuracy metrics, with little support for validating whether a prediction is correct for the right reasons. Prior work~\cite{chatzisavvas2026towards} addressed a binary segmentation task with a single-user architecture. Here we extend this to three computer vision tasks with multiclass support, a multi-user identity-federated architecture, and an interpretability loop that lets domain experts interrogate model predictions directly.

We present \emph{AmalthAI}, a web-based machine learning platform that closes both gaps at once. A unified environment with an intuitive graphical interface lets users manage datasets, configure training sessions, evaluate trained models, and perform inference without interacting directly with source code, while the platform's self-hostable, multi-user architecture keeps restricted artifact data under institutional control rather than transferring it to external services. To close the validation gap, AmalthAI integrates Grad-CAM attention maps~\cite{selvaraju2020grad}, and a vision-language model (VLM) assistant that explains individual predictions in plain language, letting domain experts judge whether the model's highlighted regions and generated descriptions align with domain-relevant visual evidence rather than relying on accuracy alone. The platform supports three application domains, image classification, semantic segmentation, and object detection, alongside configurable training options such as hyperparameter tuning and data augmentation for controlled experimentation. We demonstrate AmalthAI on a custom experimental archaeological dataset, where sample localization and classification are combined with CH expert-supported hypothesis testing to produce archaeologically informative results. This case study shows how domain experts can develop, train, and validate computer vision models while substantially reducing the technical expertise the workflow demands.

The primary contributions of this work are summarized as follows:
\begin{itemize}

\item[$\bullet$] \textbf{A Cultural Heritage-Oriented Machine Learning Platform:} An open-source platform that unifies dataset management, model training, evaluation, and inference for computer vision tasks within a single application that makes these capabilities accessible to domain experts without a machine learning background, supporting three computer vision modalities.

\item[$\bullet$] \textbf{Scalable and Modular Multi-User Architecture:} A dockerized, multi-user architecture which allows on-premise standalone deployment, integration with external identities and data services, and expansion to additional models.

\item[$\bullet$] \textbf{Experimental Validation:} The platform is evaluated on a real-world archaeological computer vision case study, demonstrating its practical effectiveness and usability across visual analysis tasks leading to meaningful findings using hypothesis testing. 

\item[$\bullet$] \textbf{Explainability, Interpretability and Expert Analysis:} A system combining Grad-CAM attention maps with VLM-generated prediction explanations, letting domain experts inspect whether model attention aligns with the archaeologically relevant region rather than relying on accuracy alone.

\end{itemize} 

\section{Related Work}

\noindent\textbf{Machine Learning Platforms.} Integrated computer vision platforms have emerged to simplify the end-to-end development lifecycle by combining dataset management, annotation, model training, evaluation, and deployment within a unified environment. Roboflow~\cite{dwyer2026roboflow} provides one of the most comprehensive ecosystems, offering dataset versioning, annotation, preprocessing, augmentation, training, and hosted inference for a wide range of vision tasks. Intel Geti~\cite{IntelGeti} emphasizes AI-assisted annotation and active learning, enabling iterative dataset refinement while supporting no-code model training, evaluation, and deployment. Another notable example is the Ultralytics Platform~\cite{UltralyticsPlatform}, which introduced a unified workflow centered on the YOLO ecosystem, integrating annotation, cloud-based training, evaluation, inference, and deployment into a streamlined user interface. While these platforms significantly reduce the engineering effort required to develop computer vision applications, they are predominantly proprietary and cloud-oriented, limiting transparency, extensibility, and self-hosted deployment. 

In cultural heritage, this limitation is particularly important because many workflows require image collections, annotations and metadata to be stored in external databases or cloud services. For restricted artifact data, this can make the use of such platforms questionable or even incompatible with institutional, legal or collection specific distribution policies. In contrast, our work presents an open-source, modular platform that offers comparable end-to-end functionality while allowing researchers and practitioners to customize, extend, and deploy the entire pipeline within their own infrastructure, thereby facilitating reproducible research and collaborative development.\newline

\noindent\textbf{Machine Learning Operations (MLOps).} MLOps have become an essential component of modern AI systems, enabling reproducible experimentation, automated training, scalable deployment, and lifecycle management of machine learning models across heterogeneous computing environments~\cite{kreuzberger2023machine}. Several open-source frameworks have been proposed to streamline these processes. MLflow~\cite{Zaharia_Accelerating_the_Machine_2018} provides a modular platform for experiment tracking, model versioning, artifact management, and deployment, making it a popular choice for reproducible machine learning workflows. Kubeflow~\cite{kubeflow2021machine} extends these capabilities by leveraging Kubernetes to orchestrate scalable machine learning pipelines, supporting distributed training, workflow automation, and resource management for production environments. In particular, Kubeflow integrates Katib, which enables automated hyperparameter optimization through search strategies such as random search, grid search, and Bayesian optimization. ClearML~\cite{clearml} has emerged as a comprehensive open-source MLOps platform that combines experiment tracking, dataset and model versioning, pipeline orchestration, remote execution, and resource management within a unified framework, making it particularly attractive for computer vision applications. 

Our platform incorporates MLOps principles to support the complete lifecycle of vision models, from experiment management and automated training to deployment and inference. By combining computer vision workflow management with scalable MLOps functionality, AmalthAI facilitates reproducible experimentation, efficient resource utilization, and streamlined deployment while remaining fully open-source and extensible.\newline

\noindent\textbf{Use Cases in Cultural Heritage.}
Computer vision and deep learning have become increasingly important in CH research, enabling the automated analysis of large image collections that would otherwise require extensive manual effort~\cite{fiorucci2020machine}. Recent studies have applied image classification, object detection, and semantic segmentation to tasks such as artifact categorization~\cite{winterbottom2022deep}, architectural style recognition~\cite{yoshimura2019deep}, material identification~\cite{pei2024recognizing}, and damage assessment~\cite{elbehairy2026comprehensive}. These methods have been successfully employed across diverse CH assets, including textiles~\cite{sha2024image}, ceramics~\cite{ling2024findings}, and archaeological monuments~\cite{ottoni2025deep}. However, most existing works focus on developing task-specific models~\cite{jing2022mobile,chen2022improved} and require substantial expertise in machine learning frameworks and programming, limiting their accessibility to domain experts. This highlights the need for integrated and user-friendly platforms that facilitate the adoption of computer vision techniques in CH research.

\begin{figure}[ht]
    \centering
    \includegraphics[width=0.99\linewidth]{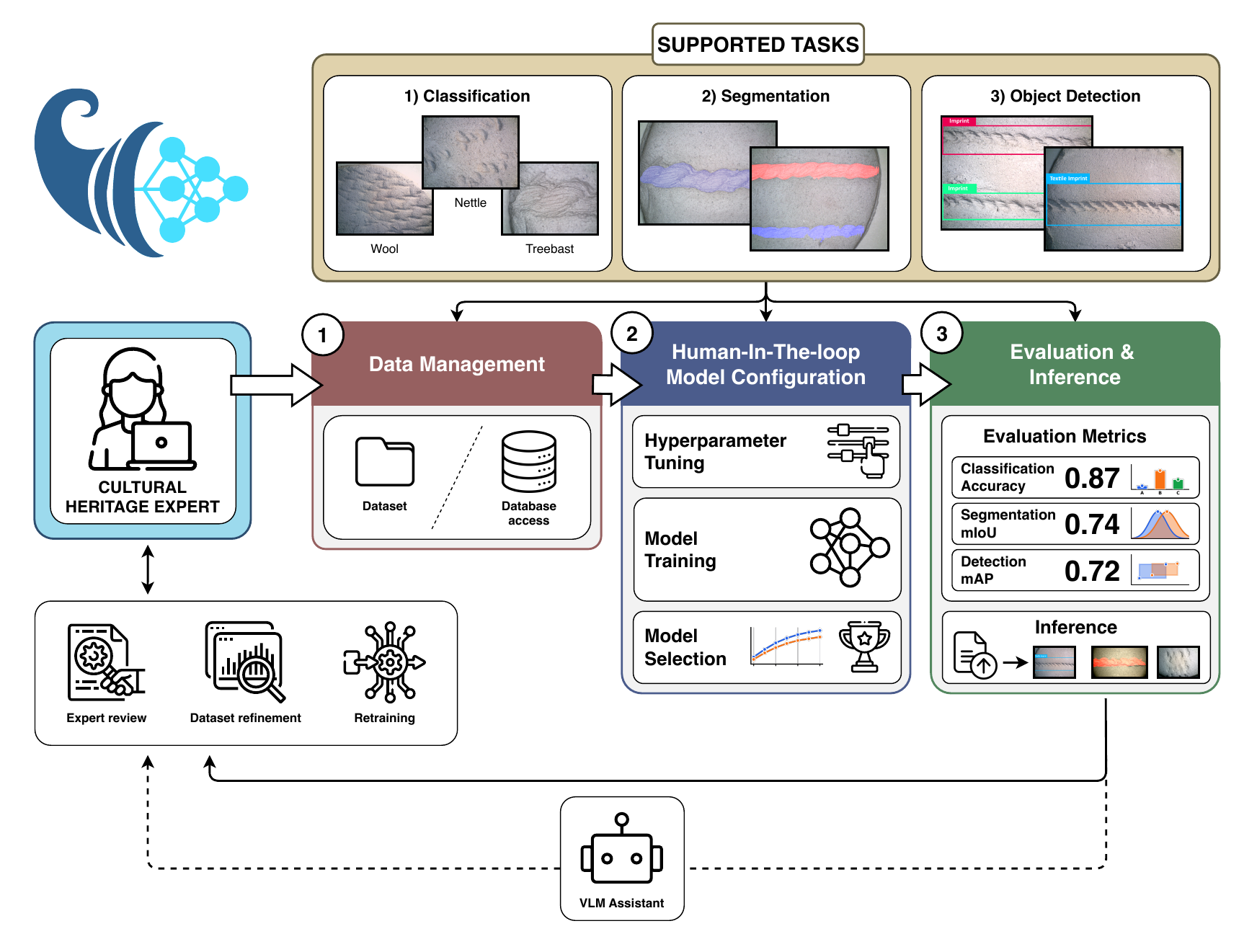}
    \caption{Overview of the AmalthAI architecture and CH computer vision workflow. The platform exposes three supported task modalities (classification, segmentation, and object detection) within a unified interface. A CH expert interacts with the system through three main stages: data management, where image datasets are imported or accessed from a database; model training, where hyperparameter tuning, training, and model selection are handled through an accessible frontend and backend execution layer; and evaluation and inference, where task-specific metrics and predictions are returned for expert review and analysis.}
    \label{fig:arch}
\end{figure}
\section{The AmalthAI Platform Architecture}

\subsection{Architectural Overview}

Following the modularity principles of the framework presented in~\cite{chatzisavvas2026towards}, AmalthAI has been developed as a fully dockerized machine learning platform that supports the complete workflow from dataset management to model deployment. Containerization ensures reproducible execution environments, simplifies deployment, and enables scalable operation across heterogeneous computing infrastructures. The platform follows a modular architecture that facilitates the integration of new computer vision tasks, machine learning models, and backend services while maintaining a consistent user experience. Fig.~\ref{fig:arch} illustrates the overall architecture of the system.

\subsection{User Interface}

The frontend is implemented using the Flask web framework~\cite{flask2025} and serves as the primary interface between users and the machine learning pipeline. Its design abstracts the underlying implementation details, allowing users to configure experiments, manage datasets, launch training jobs, and perform inference without interacting with source code or command-line tools. The configuration interface provides contextual descriptions of the available options, helping users understand the purpose of each parameter before launching an experiment. It also provides recommendations on the typical use cases and trade-offs of the available model architectures, assisting users in selecting appropriate models according to their task requirements and computational constraints. Similarly, the dataset management interface provides guidance on the expected dataset structure and organization.

\setcounter{footnote}{0}

For more detailed guidance, the platform provides direct links to the accompanying documentation\footnote{Available at \href{https://github.com/TEXTaiLES/AmalthAI-documentation}{the AmalthAI documentation}.}. The documentation contains recommendations for configuring training parameters, including suggested values and considerations for different experimental settings. It also provides dataset size requirements and guidelines regarding recommended training, validation, and test splits to assist with reliable model development. Throughout the platform, interactive visual components are used to simplify navigation and provide immediate feedback on dataset management, training progress, model evaluation, and inference results.

\subsection{MLOps \& Model Parameterization}
\label{sec3.3}

\subsubsection{Training.}

The training pipeline is built upon Kubernetes-based MLOps, with Kubeflow orchestrating the execution of training jobs and Katib performing automated hyperparameter optimization through parallel trial execution. User selections from the frontend are translated into experiment configurations that are submitted to the Kubernetes cluster for execution. In addition to automated optimization, the platform allows users to define custom search intervals for selected hyperparameters, providing greater control over the optimization process. Data augmentation policies can also be configured through the user interface, enabling comparative experiments under different preprocessing strategies.

AmalthAI currently supports three computer vision tasks: Image Classification, Semantic Segmentation, and Object Detection, with multi-class learning available across all tasks. Training from scratch is supported for all three tasks, while image classification and object detection additionally support fine-tuning of pretrained models. The training strategy can be configured directly through the user interface. For semantic segmentation, the platform integrates widely adopted architectures, including U-Net~\cite{ronneberger2015u}, DeepLabV3+~\cite{chen2018encoder}, and PSPNet~\cite{zhao2017pyramid}. Object detection is supported through the Ultralytics YOLO family of models~\cite{jocher2026ultralytics}, specifically YOLOv8, YOLO11, and YOLO26, selected for their favorable balance between accuracy, computational efficiency, and inference speed. For image classification, the platform includes several established convolutional neural network architectures, such as ResNet~\cite{he2016deep}, EfficientNet~\cite{tan2019efficientnet}, MobileNetV2~\cite{sandler2018mobilenetv2}, ShuffleNetV2~\cite{ma2018shufflenet}, and ConvNeXt~\cite{liu2022convnet}. The model repository has been designed to be easily extensible, allowing additional classification architectures available through the TorchVision library to be incorporated with minimal implementation effort.

Each training experiment may consist of multiple parallel training trials executed by Katib, where each trial evaluates a different combination of hyperparameters. Once all trials have been completed, the platform automatically selects and stores only the best-performing model checkpoint along with its associated configuration and metadata. By default, the selection criterion is the primary evaluation metric of the corresponding task: mean Intersection over Union (mIoU)~\cite{long2015fully} for Semantic Segmentation, mAP@50--95~\cite{lin2014microsoft} for Object Detection, and Accuracy~\cite{russakovsky2015imagenet} for Image Classification. However, the optimization objective is configurable and may be adapted to alternative criteria, such as minimizing the training or validation loss, depending on the experimental requirements.

\begin{figure}[ht]
    \centering
    \includegraphics[width=0.99\linewidth]{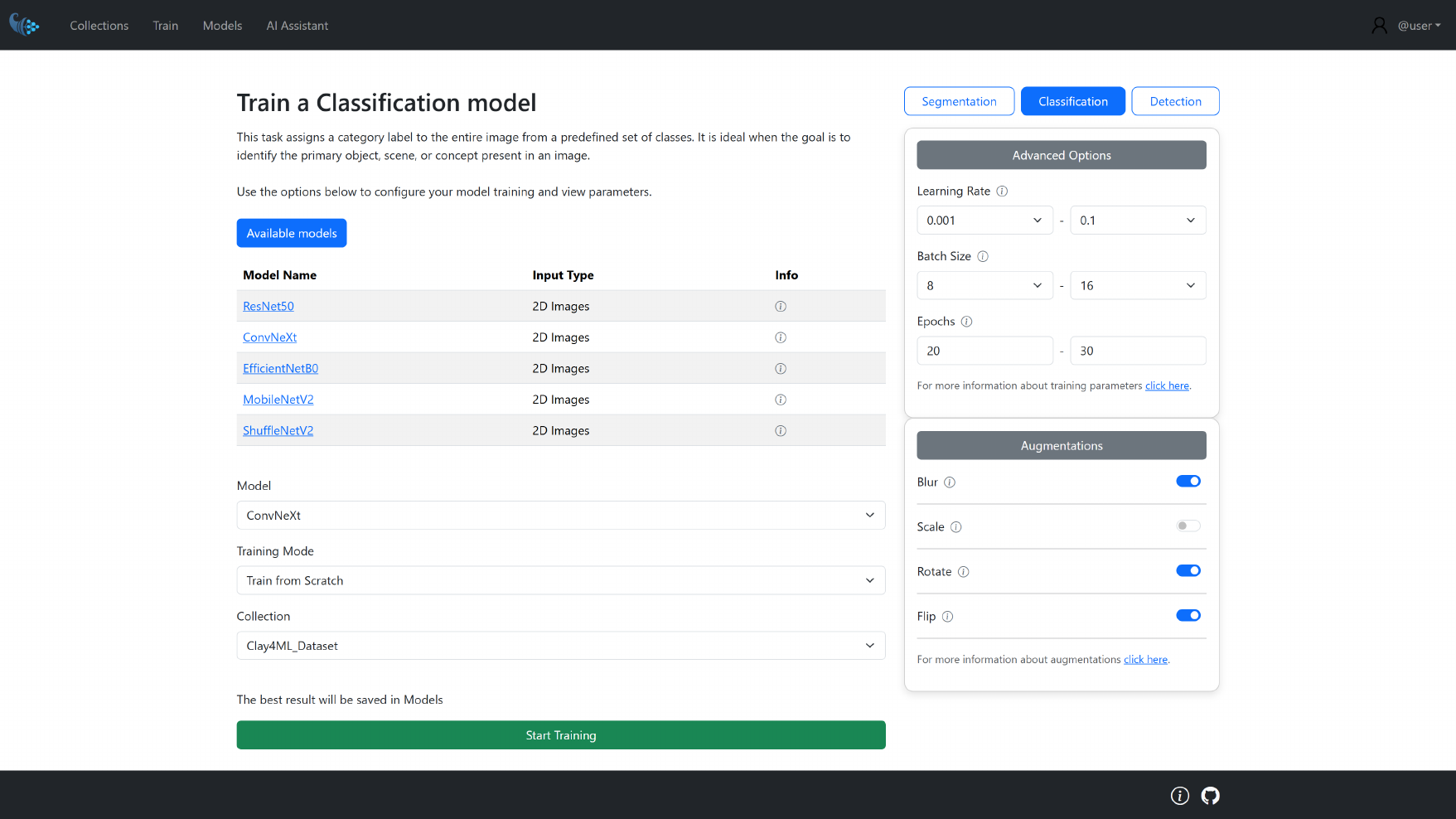}
    \caption{Model training configuration interface. The platform enables intuitive selection of model architectures, hyperparameter tuning, and data augmentations without interaction with underlying code.}
    \label{fig:train}
\end{figure}

\subsubsection{Inference.}

\begin{figure}[t]
  \centering
  \includegraphics[width=1.00\linewidth]{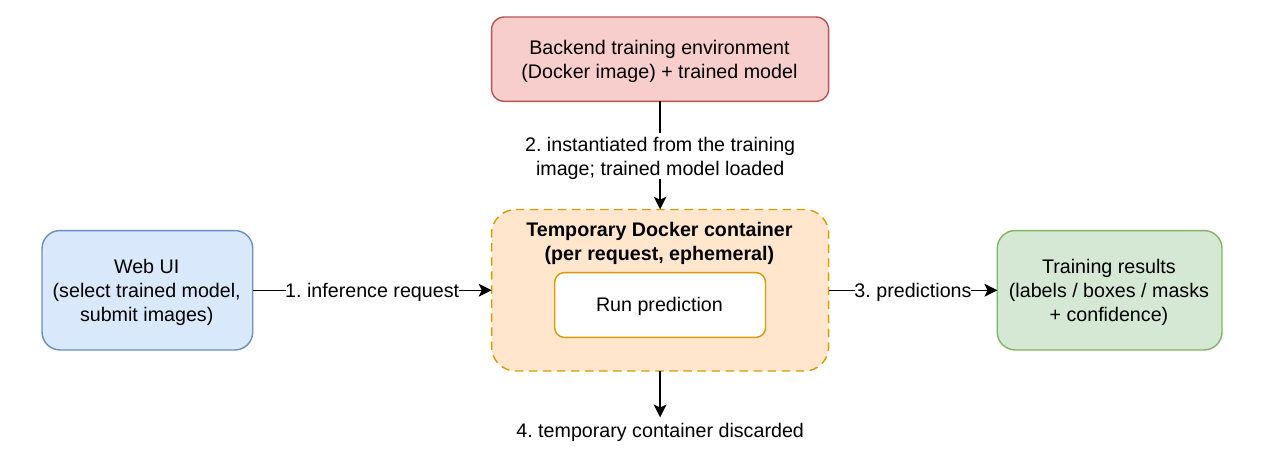}
  \caption{Stateless inference. For each request, the platform instantiates a temporary Docker container from the same software environment used during training, loads the selected trained model, performs the prediction, returns the results to the interface, and discards the container on completion.}
  \label{fig:inference-container}
\end{figure}

The inference module allows users to deploy previously trained models directly through the web interface. Users may submit either a single image or multiple images simultaneously for batch inference. For every inference request, the platform instantiates a temporary Docker container using the same software environment employed during training. The container performs the requested predictions and is automatically discarded upon completion (Fig.~\ref{fig:inference-container}). This stateless execution model ensures consistency between training and inference, while maintaining efficient utilization of computational resources. Depending on the selected task, the platform returns image classification predictions, object detections, or semantic segmentation masks, together with their corresponding confidence scores. The prediction results are presented alongside the original input images through the web interface. Additionally, for the image classification task and to support model explainability, the platform also returns Grad-CAM~\cite{selvaraju2020grad} visualizations alongside the predicted class probabilities. These heatmaps highlight the regions of the input image that most strongly influenced the model’s decision, enabling users to better understand which visual features the model focused on when producing its prediction.

\subsection{Vision-Language Assistant}

In addition to its machine learning capabilities, AmalthAI integrates a Vision-Language Model (VLM), specifically Qwen2-VL-2B-Instruct~\cite{wang2024qwen2}, to support explainable decision-making. The VLM takes as input both artifact images and the corresponding Grad-CAM~\cite{selvaraju2020grad} visualizations to generate natural-language explanations of classifier decisions, particularly in cases of misclassification. Rather than replacing domain expertise, it functions as a decision-support tool, providing evidence-based observations, suggesting plausible interpretations, and indicating uncertainty when visual evidence is insufficient. Representative visual examples together with the corresponding prompts are included in the supplementary material. The pipeline is designed in a way that requires domain knowledge rather than ML or VLM engineering expertise. Users are not expected to manually design system prompts; AmalthAI maintains templates while the user provides a small set of dataset specific knowledge during configuration which is then used by the platform to adapt the prompt.

\subsection{User Management}

AmalthAI supports concurrent, multi-user operation within a shared deployment. In a standalone setting, user authentication is managed by the platform itself. Alternatively, authentication can be delegated to a third-party identity provider, making AmalthAI compatible with established identity and access-management systems, such as Directus~\cite{directus}, Keycloak~\cite{keycloak}, or Authentik~\cite{authentik}, and, more broadly, with any service implementing the OAuth~2.0~\cite{rfc6749} and OpenID~Connect~\cite{openidconnect} standards. This enables seamless integration with organizational single sign-on and easy interconnection with other services.

In both cases, AmalthAI associates authenticated users with a stable, unique identifier, valid across every resource and reference they create within the system. Therefore, all uploaded datasets, training experiments, model checkpoints and inference results are isolated per user, while the underlying computational and storage resources remain shared. This ensures a clean, privacy-respecting workflow, while enabling effective collaboration and organization across users.

Since authentication is decoupled from the rest of the platform, and per-user identifiers are persistent across the system, local user identities can be trivially reconciled with external accounts when such a service is used, contributing to the modular nature of the platform. 

\subsection{Data Storage \& Deployment}

Data management follows a similarly modular pattern: the entire lifecycle of all user assets can be self-managed by the tool or delegated to an external data service through an abstracted storage layer that provides data persistence and external service connectivity.

\begin{figure}[t]
  \centering
  \includegraphics[width=\linewidth]{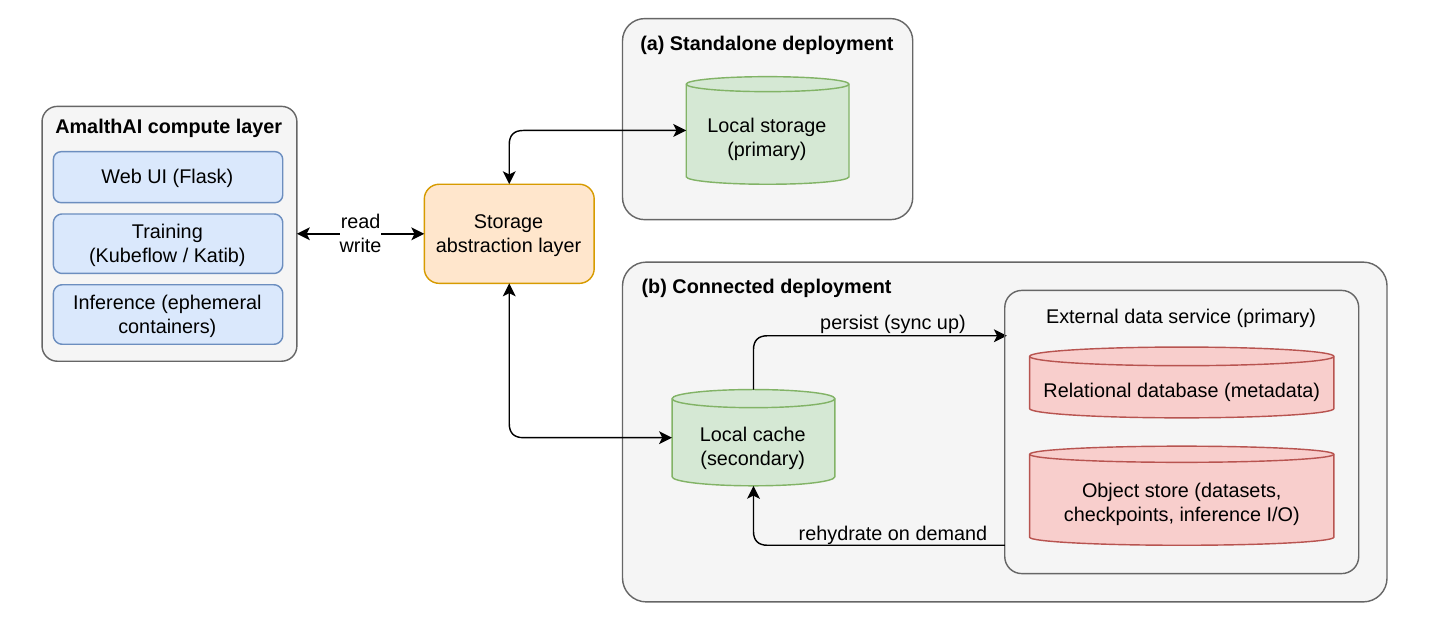}
  \caption{AmalthAI's modular data layer. (a) Standalone: assets are read from and written to local storage through the storage-abstraction layer; no external service is required. (b) With an external service, the local directory becomes a cache: assets are rehydrated on first use and new ones pushed back in two-way, content-addressed synchronization, with automatic fallback to the cache when the service is offline.}
  \label{fig:data-layer}
\end{figure}

By default, the platform is fully self-contained: datasets, training experiments and configurations, trained model checkpoints and inference inputs/outputs are persisted to local storage on the machine AmalthAI is deployed on, allowing a completely standalone deployment without external dependencies (Fig.~\ref{fig:data-layer}a). This is particularly important for archaeological and cultural heritage datasets that cannot be transferred to external databases or cloud services due to ownership or distribution restrictions.

At the same time, the tool's architecture supports complete asset management via an external service when a larger infrastructure is needed. When an external data service is configured (Fig.~\ref{fig:data-layer}b), AmalthAI treats its local working directories as a cache over that service rather than as the primary copy, retrieving assets required for a training or inference job on demand into local storage. Transfers are content-addressed to avoid redundant data movement, and a local index of all available assets is maintained and synchronized both ways with the remote service. In case the external service is momentarily unavailable, the system falls back to the locally cached state, and automatically re-synchronizes the moment it re-establishes a connection to the network.

This yields the benefits of decoupled, synchronized storage management: end users and infrastructure operators may adopt any external storage service, or none at all, while still obtaining two-way synchronization, durable data persistence and automatic caching with local fallback.

The platform is distributed as a collection of Docker images, ensuring reproducible deployment across heterogeneous infrastructures. In a standalone configuration, AmalthAI runs as a self-sufficient unit on a single host. In a larger deployment, the compute layer and the data storage service operate as independent, separately configurable units that communicate over the network. Scalable training and inference are handled by the Kubernetes-in-Docker cluster, with data management able to be configured in several ways (\eg a relational database for structured metadata combined with an S3-compatible object store for large binary data, connected via REST API), or left to the default local storage solution. This separation enables AmalthAI to scale from a single-machine installation to a component of a shared, multi-platform ecosystem with minimal configuration.
\section{Workflow Validation Through an Archaeological Case Study}
\label{sec:archaeological-use-case}

\begin{figure}[ht]
    \centering
    \includegraphics[width=0.75\linewidth]{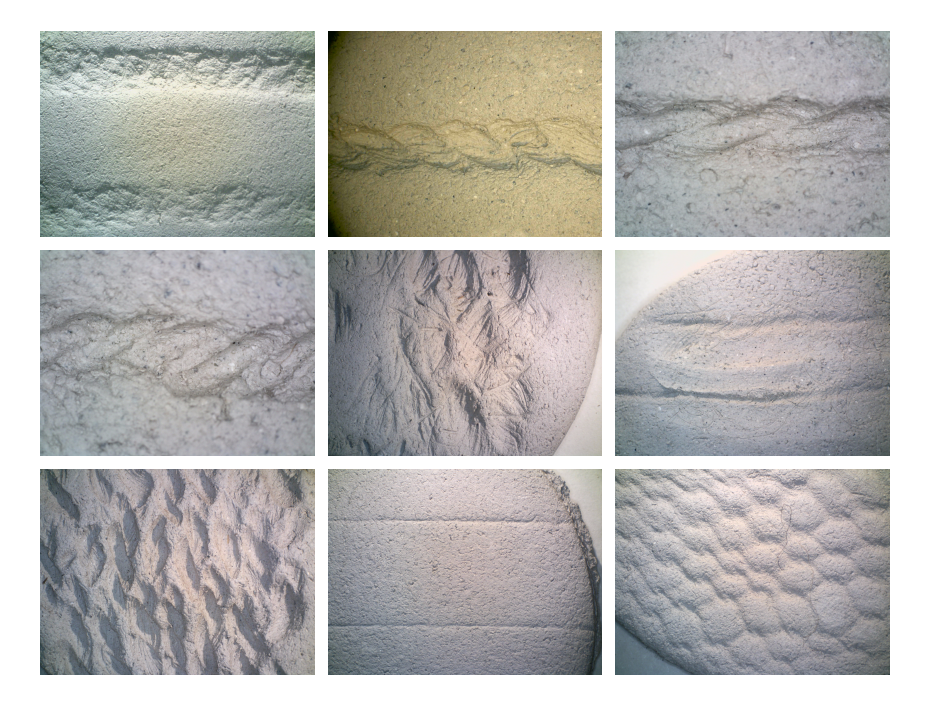}
    \caption{Indicatory samples from the experimental textile imprint dataset. Samples have been selected to represent the variety and complexity of the dataset across different clay textures, lighting conditions and camera settings.}
    \label{fig:imprints}
\end{figure}

Experimental archaeology and machine learning have been intertwined in research on cereal growing conditions \cite{ORENGO2026106607}, use-wear analysis \cite{Eleftheriadou-2025, sferrazza2025archaeological}, lithic microdebitage \cite{eberl2023machine}, and can be considered a promising direction in the studies of archaeological textile imprints. The approach offers possibility of filling the gap of the limited  amount of archaeological remains of the textile imprints by creating data in the controlled conditions. To ground AmalthAI in a realistic cultural heritage workflow, we use an experimental archaeology case study concerning textile imprints on clay. The relevance of such imprints lies in the indirect evidence that may be left of the original textile, such as raw materials, textile structure, and production techniques. The dataset was produced from controlled textile samples impressed on clay and documented through close-range and microscopic imaging (Dino-Lite) and was curated and annotated by CH experts. Indicatory samples of the experimental dataset can be seen in Fig.~\ref{fig:imprints}.

\subsection{Hypothesis Testing in Archaeology}

Hypothesis testing refers to the process of evaluating an expert-formulated proposition through observable evidence~\cite{10.1098/rsta.1933.0009, outram}. In this case study, the central question regarding the examined use case is the following: \textit{Can clay imprints preserve archaeologically meaningful visual information from the original textile, and, if so, can this information be recovered through computer vision models?} This question is particularly relevant in textile archaeology, where preserved textile artifacts are uncommon, fragile, and often difficult to examine directly~\cite{cybulska}.

The hypothesis regarding this case study is formulated as follows: textile imprints on clay preserve visual features that are informative of the original textile's raw material and production technique, and these features can be exploited by machine learning models trained on controlled experimental data. The hypothesis is considered supported if the model can effectively classify the experimental data according to these attributes and that misclassifications can be attributed to the structural similarities between the corresponding textile classes.

\subsection{Use Case Workflow and Task Formulation}

As a computer vision task, the use case can be viewed as a binary segmentation problem, with the purpose of precisely isolating the area of the textile imprint, and two classification problems, where each image can be classified according to production technique and raw material. The experiments for both tasks were conducted by CH experts with minimal technical knowledge in machine learning. No expert guidance was provided directly throughout the process besides the platform's documentation and implemented information. The final results were evaluated by both machine learning and CH experts to confirm expected model behavior and interpretable results.\newline

\noindent\textbf{Experimental Setup.}
For classification, the dataset was split into classes according to production technique and raw material before being uploaded and processed by AmalthAI, resulting in 1927 images for material classification and 1837 images for technique classification, with only a few samples outside the defined classes. For segmentation, a curated subset of 387 images was used, with masks manually annotated by CH experts using the CVAT annotator~\cite{CVAT_ai_Corporation_Computer_Vision_Annotation_2023} integrated with SAM~\cite{kirillov2023segment}. Across all tasks, training was configured to include all models available for the corresponding modality, while task-specific hyperparameter ranges and augmentations were selected through the platform, as summarized in Tab.~\ref{tab:textile-imprint-training-config}. The best-performing models were saved and subsequently used for inference. Given the experimental nature of the data, the training process was repeated with different training/test splits, with all splits producing similar results, as documented in Tab.~\ref{tab:textile-imprint-results}.

\begin{table}[ht]
    \centering
    \caption{Training configuration used for the textile-imprint classification and segmentation experiments.}
    \label{tab:textile-imprint-training-config}
    \setlength{\tabcolsep}{10pt}
    \begin{tabular}{l c c c c}
        \hline
        Task & Dataset size & Learning rate & Batch size & Epochs \\
        \hline
        Classification & 1927 / 1837 & $0.001$--$0.1$ & $4$--$16$ & $30$--$50$ \\
        Segmentation & 387 & $0.001$--$0.1$ & $8$--$16$ & $20$--$30$ \\
        \hline
    \end{tabular}
\end{table}

\begin{table}[t]
    \centering
    \caption{Performance results of the AmalthAI workflow on the experimental textile-imprint use case. Classification is reported as accuracy for material and production-technique prediction, while segmentation is reported as mIoU for imprint-region delineation. Values are reported as mean $\pm$ standard deviation over 5 experimental splits.}
    \label{tab:textile-imprint-results}

    \begin{subtable}{0.85\linewidth}
        \centering
        \caption{Classification results.}
        \label{tab:textile-imprint-classification-results}
        \begin{tabular}{l @{\hspace{1.6em}} c @{\hspace{1.6em}} c}
            \hline
            Model & Material Acc. (\%) & Technique Acc. (\%) \\
            \hline
            ResNet18~\cite{he2016deep} & $74.88 \pm 1.04$ & $83.09 \pm 2.09$ \\
            EfficientNetB0~\cite{tan2019efficientnet} & \textbf{\boldmath{$76.32 \pm 1.21$}} & \textbf{\boldmath{$83.68 \pm 1.89$}} \\
            MobileNetV2~\cite{sandler2018mobilenetv2} & $74.32 \pm 1.33$ & $82.89 \pm 2.37$ \\
            ShuffleNetV2~\cite{ma2018shufflenet} & $72.28 \pm 1.98$ & $80.10 \pm 2.66$ \\
            \hline
        \end{tabular}
    \end{subtable}

    \vspace{0.6em}

    \begin{subtable}{0.55\linewidth}
        \centering
        \caption{Segmentation results.}
        \label{tab:textile-imprint-segmentation-results}
        \begin{tabular}{l @{\hspace{2em}} c}
            \hline
            Model & mIoU (\%) \\
            \hline
            UNet~\cite{ronneberger2015u} & $90.02 \pm 1.79$ \\
            DeeplabV3+~\cite{chen2018encoder} & \textbf{\boldmath{$90.47 \pm 1.78$}} \\
            PSPNet~\cite{zhao2017pyramid} & $84.20 \pm 3.60$ \\
            \hline
        \end{tabular}
    \end{subtable}
\end{table}

\subsection{Results and Workflow Validation}
The results displayed in Tab.~\ref{tab:textile-imprint-results}, together with feedback from the end users, suggest that the workflow is practical for standard computer vision tasks in an archaeological setting. According to the CH experts, the resulting scores, as well as inference visualizations, are consistent with the properties, size and visual complexity of the dataset. The main observation is operational: AmalthAI allows a CH expert to move from a structured image dataset to a trained, reusable model without directly managing code, dependencies or backend execution. The experiments also illustrate the value of a reproducible workflow for cultural heritage data analysis. Since the samples, imaging conditions, labels, training configuration, and trained models can be organized within a single platform, the resulting workflow is easier to inspect, repeat, and compare across dataset splits or task formulations. We note that the provided model results are intended as workflow-validation evidence rather than as a benchmark claim. \newline

\noindent\textbf{Classification results.}
The aforementioned classification experiments demonstrate that AmalthAI can support parallel evaluation of multiple standard image classification architectures. The repeated train/test splits produced similar scores, suggesting that the observed performance was not dependent on a single favorable data partition. From a workflow perspective, the important result is that the platform can configure, execute, compare, register, and reuse classification models for more than one archaeological label set. \newline

\noindent\textbf{Segmentation results.}
The segmentation task has a different role from the classification experiments. While it does not directly infer raw material or production technique, it provides a precise localization layer for the imprint region. This is important because the textile imprint images contain substantial non-diagnostic visual information, including clay texture, lighting variation, and background surface irregularities. A reliable segmentation model therefore supports a more controlled analytical workflow by separating the impressed textile trace from its surrounding matrix. The numerical results indicate that this localization step can be reproduced consistently, despite the limited number of manually annotated masks. A set of qualitative segmentation results can be viewed in the supplementary material. 

\subsubsection{Hypothesis-testing evidence.}
The classification inference results in Tab.~\ref{tab:classification_results} provide evidence relevant to the hypothesis formulated in Sec.~\ref{sec:archaeological-use-case}. The trained models classify the textile-imprint samples according to both raw material and production technique with meaningful overall performance, reaching $77.53\%$ accuracy for material classification and $85.57\%$ accuracy for technique classification. These results indicate that the imprints contain recoverable visual information associated with the archaeological labels.

The class-level results are also informative from a cultural heritage perspective. Misclassifications are not distributed uniformly across all classes: nettle shows lower material classification accuracy, while splicing is the most difficult production-technique class. According to CH experts, these are also among the cases that are more difficult to distinguish through visual inspection, especially when fibres have been processed in similar ways~\cite{ulanowskaWhyNotWool2023}. This correspondence is consistent with the models exploiting label relevant visual properties, although it does not by itself exclude reliance on dataset specific cues.

Taken together, the results are consistent with the hypothesis that experimentally produced clay imprints can preserve label-relevant information from textile samples. They do not imply that every imprint preserves the same level of information as the original textile, nor that automated classification can replace expert analysis. Rather, within this controlled dataset, the results suggest that textile-imprint imagery can support computational analysis of raw material and production technique, and that AmalthAI can provide an accessible workflow for testing such archaeological propositions.

\begin{table}[ht]
    \centering
    \caption{Inference results from the classification tasks used for hypothesis testing. The models used for the experiment were the best performing models from each task (EfficientNetB0~\cite{tan2019efficientnet}) as documented in Tab.~\ref{tab:textile-imprint-classification-results}.}
    \label{tab:classification_results}
    \begin{tabular}{l c c @{\hspace{2.2em}} l c c}
        \hline
        \multicolumn{3}{c}{Material} & \multicolumn{3}{c}{Technique} \\
        \hline
        Class & Correct / Total & Acc. (\%) & Class & Correct / Total & Acc. (\%) \\
        \hline
        Flax & 195 / 267 & 73.03 & Drilling & 644 / 727 & 88.58 \\
        Nettle & 385 / 560 & 68.75 & Spinning & 691 / 761 & 90.80 \\
        Lime Bast & 433 / 550 & 78.73 & Splicing & 237 / 349 & 67.91 \\
        Wool & 481 / 550 & 87.45 &  &  &  \\
        Overall & 1494 / 1927 & 77.53 & Overall & 1572 / 1837 & 85.57 \\
        \hline
    \end{tabular}
\end{table}

\subsubsection{Limitations and Future Work.}

AmalthAI is designed to make computer vision workflows more accessible to cultural heritage experts by providing a streamlined interface for dataset handling, model training, evaluation, and inference. While this design reduces the amount of technical configuration required from the user, it also limits the degree of control available through the interface. Although the platform exposes several commonly used training parameters, more advanced options, such as custom loss functions, optimizer selection, scheduler configuration, and other low-level training settings, are currently not accessible to end users. Furthermore, AmalthAI currently requires all datasets to be strictly pre-partitioned by the user prior to upload, without internal validation of the data distribution.

Future work will focus on extending the platform's functionality while preserving its ease of use. To address the vulnerability of non-expert users to overfitting or poor generalization due to inadequate data preparation, automated data sanity checks will be incorporated during the upload phase. This update will provide in-app guidance to flag small dataset sizes, warn against severe class imbalances, and evaluate user-provided split ratios before training begins. Another promising direction is the introduction of role-based access control, enabling advanced users to access more sophisticated training configurations without increasing complexity for non-expert users. Finally, the current platform is limited to 2D RGB imagery. Therefore, an important enhancement is the support for multispectral imagery, as spectral bands beyond the visible range can reveal material properties and features that are not detectable in conventional RGB images, providing richer and more informative input for AI models~\cite{Liang2011AdvancesIM}.
\section{Conclusion}
\label{sec:conclusion}

We presented AmalthAI, an open-source computer vision platform designed for use in cultural heritage machine learning workflows. The standard machine learning pipeline, which includes data handling, task selection, training configuration, and inference, is served as a unified workflow through an intuitive web interface and contextual guidance, making it approachable to experts unfamiliar with machine learning. The platform's usability has been validated through a real-world case study, establishing it as a useful tool for effective data analysis in the field. By supporting three major computer vision tasks, individual user management and modular data storage, hostable on both internal and external premises, it ensures a practical and scalable bridge between established computer vision methods and cultural heritage research.
\section*{Acknowledgments}
This work has been partially supported by the European Union’s Horizon Europe research and innovation programme under Grant Agreement No. 101158328 (TEXTaiLES), and by the European High-Performance Computing Joint Undertaking (JU) under Grant Agreement No. 101234269 for the Pharos AI Factory project, as well as by the Greek Ministry of Digital Governance and Artificial Intelligence. Additional support was provided by project MIS 5154714 of the National Recovery and Resilience Plan Greece 2.0, funded by the European Union under the NextGenerationEU Programme.

The dataset was produced by archaeologists from the Faculty of Archaeology, University of Warsaw, involved in the Textile Digitisation Tools and Methods for Cultural Heritage (TEXTaiLES) and Exploring Textile Imprints on Clay from the 3rd and the 2nd Millennia BCE: Advancing Cutting-Edge Research and Documentation Protocols with Case Studies of Diverse Textile Consumption Contexts (ExplorTIC) projects, in close collaboration with the Biskupin Archaeological Museum, Poland.


%
%
\bibliographystyle{splncs04}
\bibliography{main}

@String(CVPR  = {IEEE Conf. Comput. Vis. Pattern Recog.})

@String(ECCV  = {Eur. Conf. Comput. Vis.})

@String(ICPR  = {Int. Conf. Pattern Recog.})

@String(CVPR  = {CVPR})

@String(ECCV  = {ECCV})

@String(ICPR  = {ICPR})

@inproceedings{chatzisavvas2026towards,
  author    = {Christos Chatzisavvas and Thomas Pappas and Panagiotis Rigas and Nikolaos Mitianoudis and George Pavlidis and Chairi Kiourt and Anestis Koutsoudis and Vassilis Katsouros and George Ioannakis},
  title     = {Towards an Easy-to-Use Machine Learning Framework for Cultural Heritage Scientists},
  booktitle = {Computer Applications and Quantitative Methods in Archaeology Conference (CAA2025)},
  address   = {Athens, Greece},
  year      = {2026},
  month     = may,
  day       = {6},
  doi       = {10.5281/zenodo.20048428}
}

@inproceedings{ronneberger2015u,
  title={U-net: Convolutional networks for biomedical image segmentation},
  author={Ronneberger, Olaf and Fischer, Philipp and Brox, Thomas},
  booktitle={International Conference on Medical image computing and computer-assisted intervention},
  pages={234--241},
  year={2015},
  organization={Springer}
}

@inproceedings{zhao2017pyramid,
  title={Pyramid scene parsing network},
  author={Zhao, Hengshuang and Shi, Jianping and Qi, Xiaojuan and Wang, Xiaogang and Jia, Jiaya},
  booktitle={Proceedings of the IEEE conference on computer vision and pattern recognition},
  pages={2881--2890},
  year={2017}
}

@inproceedings{chen2018encoder,
  title={Encoder-decoder with atrous separable convolution for semantic image segmentation},
  author={Chen, Liang-Chieh and Zhu, Yukun and Papandreou, George and Schroff, Florian and Adam, Hartwig},
  booktitle={Proceedings of the European conference on computer vision (ECCV)},
  pages={801--818},
  year={2018}
}

@inproceedings{he2016deep,
  title={Deep residual learning for image recognition},
  author={He, Kaiming and Zhang, Xiangyu and Ren, Shaoqing and Sun, Jian},
  booktitle={Proceedings of the IEEE conference on computer vision and pattern recognition},
  pages={770--778},
  year={2016}
}

@inproceedings{tan2019efficientnet,
  title={Efficientnet: Rethinking model scaling for convolutional neural networks},
  author={Tan, Mingxing and Le, Quoc},
  booktitle={International conference on machine learning},
  pages={6105--6114},
  year={2019},
  organization={PMLR}
}

@inproceedings{sandler2018mobilenetv2,
  title={Mobilenetv2: Inverted residuals and linear bottlenecks},
  author={Sandler, Mark and Howard, Andrew and Zhu, Menglong and Zhmoginov, Andrey and Chen, Liang-Chieh},
  booktitle={Proceedings of the IEEE conference on computer vision and pattern recognition},
  pages={4510--4520},
  year={2018}
}

@inproceedings{ma2018shufflenet,
  title={Shufflenet v2: Practical guidelines for efficient {CNN} architecture design},
  author={Ma, Ningning and Zhang, Xiangyu and Zheng, Hai-Tao and Sun, Jian},
  booktitle={Proceedings of the European conference on computer vision (ECCV)},
  pages={116--131},
  year={2018}
}

@article{jocher2026ultralytics,
  title={{Ultralytics YOLO26: Unified Real-Time End-to-End Vision Models}},
  author={Jocher, Glenn and Qiu, Jing and Liu, Mengyu and Lyu, Shuai and Akyon, Fatih Cagatay and Kalfaoglu, Muhammet Esat},
  journal={arXiv preprint arXiv:2606.03748},
  year={2026}
}

@article{kreuzberger2023machine,
  title={Machine learning operations ({MLops}): Overview, definition, and architecture},
  author={Kreuzberger, Dominik and K{\"u}hl, Niklas and Hirschl, Sebastian},
  journal={IEEE access},
  volume={11},
  pages={31866--31879},
  year={2023},
  publisher={IEEE}
}

@article{Zaharia_Accelerating_the_Machine_2018,
author = {Zaharia, Matei A. and Chen, Andrew and Davidson, Aaron and Ghodsi, Ali and Hong, Sue Ann and Konwinski, Andy and Murching, Siddharth and Nykodym, Tomas and Ogilvie, Paul and Parkhe, Mani and Xie, Fen and Zumar, Corey},
journal = {IEEE Data Eng. Bull.},
pages = {39--45},
title = {{Accelerating the Machine Learning Lifecycle with MLflow}},
volume = {41},
year = {2018}
}

@misc{kubeflow2021machine,
  title={A machine learning toolkit for {Kubernetes}},
  author={Kubeflow},
  year={2021},
  note = {Accessed: Jun. 28, 2026},
  url = {https://www.kubeflow.org/}
}

@misc{clearml,
title = {ClearML - Your entire MLOps stack in one open-source tool},
year = {2024},
url={https://clear.ml/},
note = {Accessed: Jun. 28, 2026},
author = {ClearML},
}

@article{paszke2019pytorch,
  title={Pytorch: An imperative style, high-performance deep learning library},
  author={Paszke, Adam and Gross, Sam and Massa, Francisco and Lerer, Adam and Bradbury, James and Chanan, Gregory and Killeen, Trevor and Lin, Zeming and Gimelshein, Natalia and Antiga, Luca and others},
  journal={Advances in neural information processing systems},
  volume={32},
  year={2019}
}

@inproceedings{abadi2016tensorflow,
  title={{TensorFlow}: a system for {Large-Scale} machine learning},
  author={Abadi, Mart{\'\i}n and Barham, Paul and Chen, Jianmin and Chen, Zhifeng and Davis, Andy and Dean, Jeffrey and Devin, Matthieu and Ghemawat, Sanjay and Irving, Geoffrey and Isard, Michael and others},
  booktitle={12th USENIX symposium on operating systems design and implementation (OSDI 16)},
  pages={265--283},
  year={2016}
}

@article{rei2024multimodal,
  title={Multimodal metadata assignment for cultural heritage artifacts},
  author={Rei, Luis and Mladeni{\'c}, Dunja and Dorozynski, Mareike and Rottensteiner, Franz and Schleider, Thomas and Troncy, Rapha{\"e}l and Lozano, Jorge Sebasti{\'a}n and Salvatella, Mar Gait{\'a}n},
  journal={arXiv preprint arXiv:2406.00423},
  year={2024}
}

@article{ottoni2025deep,
  title={A deep learning approach for cultural heritage building classification using transfer learning and data augmentation},
  author={Ottoni, Andre Luiz Carvalho and Ottoni, Lara Toledo Cordeiro},
  journal={Journal of Cultural Heritage},
  volume={74},
  pages={214--224},
  year={2025},
  publisher={Elsevier}
}

@inproceedings{ragusa2021semantic,
  title={Semantic object segmentation in cultural sites using real and synthetic data},
  author={Ragusa, Francesco and Di Mauro, Daniele and Palermo, Alfio and Furnari, Antonino and Farinella, Giovanni Maria},
  booktitle={2020 25th International Conference on Pattern Recognition (ICPR)},
  pages={1964--1971},
  year={2021},
  organization={IEEE}
}

@misc{IntelGeti,
  author = {Intel},
  title = {{Intel Geti}},
  url = {https://github.com/open-edge-platform/geti},
  note = {Accessed: Jun. 25, 2026},
  year={2023},
}

@misc{UltralyticsPlatform,
  author = {Ultralytics},
  title = {{Ultralytics Platform}},
  url = {https://platform.ultralytics.com/},
  note = {Accessed: Jun. 25, 2026},
  year={2026},
}

@misc{dwyer2026roboflow,
  author = {Dwyer, B. and Nelson, J. and Hansen, T. and others},
  title  = {Roboflow (Version 1.0)},
  year   = {2026},
  note   = {Accessed: Jun. 25, 2026},
  url    = {https://roboflow.com},
}

@misc{flask2025,
  author = {Pallets},
  title = {Flask (Version 3.1.0)},
  year = {2024},
  note         = {Accessed: Jun. 28, 2026},
  howpublished = {https://palletsprojects.com},
}

@misc{directus,
  author       = {{Directus}},
  title        = {{Directus}: The Collaborative Backend and Headless CMS},
  howpublished = {\url{https://directus.com}},
  note         = {Accessed: Jun. 29, 2026},
  year         = {2026},
}

@misc{keycloak,
  author       = {{Keycloak}},
  title        = {{Keycloak}: Open Source Identity and Access Management},
  howpublished = {\url{https://www.keycloak.org}},
  note         = {Accessed: Jun. 28, 2026},
  year         = {2026},
}

@misc{authentik,
  author       = {{Authentik Security Inc.}},
  title        = {authentik: The authentication glue you need.},
  howpublished = {\url{https://goauthentik.io}},
  note         = {Accessed: Jun. 28, 2026},
  year         = {2026},
}

@misc{rfc6749,
    series =    {Request for Comments},
    number =    6749,
    howpublished =  {RFC 6749},
    publisher = {RFC Editor},
    doi =       {10.17487/RFC6749},
    url =       {https://www.rfc-editor.org/info/rfc6749/},
    author =    {D. Hardt},
    title =     {{The OAuth 2.0 Authorization Framework}},
    pagetotal = 76,
    year =      2012,
    month =     oct,
}

@article{openidconnect,
  title={OpenID Connect Core 1.0 incorporating errata set 1},
  author={Sakimura, Nat and Bradley, John and Jones, Mike and De Medeiros, Breno and Mortimore, Chuck},
  journal={The OpenID Foundation, specification},
  volume={335},
  year={2014}
}

@article{selvaraju2020grad,
  title={Grad-CAM: visual explanations from deep networks via gradient-based localization},
  author={Selvaraju, Ramprasaath R and Cogswell, Michael and Das, Abhishek and Vedantam, Ramakrishna and Parikh, Devi and Batra, Dhruv},
  journal={International journal of computer vision},
  volume={128},
  number={2},
  pages={336--359},
  year={2020},
  publisher={Springer}
}

@article{Liang2011AdvancesIM,
  title={Advances in multispectral and hyperspectral imaging for archaeology and art conservation},
  author={Haida Liang},
  journal={Applied Physics A},
  year={2011},
  volume={106},
  pages={309 - 323},
}

@article{fiorucci2020machine,
  title={Machine learning for cultural heritage: A survey},
  author={Fiorucci, Marco and Khoroshiltseva, Marina and Pontil, Massimiliano and Traviglia, Arianna and Del Bue, Alessio and James, Stuart},
  journal={Pattern Recognition Letters},
  volume={133},
  pages={102--108},
  year={2020},
  publisher={Elsevier}
}

@article{jing2022mobile,
  title={{Mobile-Unet}: An efficient convolutional neural network for fabric defect detection},
  author={Jing, Junfeng and Wang, Zhen and R{\"a}tsch, Matthias and Zhang, Huanhuan},
  journal={Textile research journal},
  volume={92},
  number={1-2},
  pages={30--42},
  year={2022},
  publisher={SAGE Publications Sage UK: London, England}
}

@article{chen2022improved,
  title={Improved faster {R-CNN} for fabric defect detection based on {Gabor} filter with {Genetic Algorithm optimization}},
  author={Chen, Mengqi and Yu, Lingjie and Zhi, Chao and Sun, Runjun and Zhu, Shuangwu and Gao, Zhongyuan and Ke, Zhenxia and Zhu, Mengqiu and Zhang, Yuming},
  journal={Computers in Industry},
  volume={134},
  pages={103551},
  year={2022},
  publisher={Elsevier}
}

@inproceedings{kirillov2023segment,
  title={Segment anything},
  author={Kirillov, Alexander and Mintun, Eric and Ravi, Nikhila and Mao, Hanzi and Rolland, Chloe and Gustafson, Laura and Xiao, Tete and Whitehead, Spencer and Berg, Alexander C and Lo, Wan-Yen and others},
  booktitle={Proceedings of the IEEE/CVF international conference on computer vision},
  pages={4015--4026},
  year={2023}
}

@misc{CVAT_ai_Corporation_Computer_Vision_Annotation_2023,
author = {{CVAT.ai Corporation}},
license = {MIT},
month = nov,
title = {{Computer Vision Annotation Tool (CVAT)}},
url = {https://github.com/cvat-ai/cvat},
version = {2.25.0},
note = {Accessed: Jun. 28, 2026},
year = {2023}
}

@article{ling2024findings,
  title={Findings on machine learning for identification of archaeological ceramics: A systematic literature review},
  author={Ling, Ziyao and Delnevo, Giovanni and Salomoni, Paola and Mirri, Silvia},
  journal={IEEE Access},
  volume={12},
  pages={100167--100185},
  year={2024},
  publisher={IEEE}
}

@article{sha2024image,
  title={Image classification and restoration of ancient textiles based on convolutional neural network},
  author={Sha, Sha and Li, Yi and Wei, Wantong and Liu, Yating and Chi, Cheng and Jiang, Xuewei and Deng, Zhongmin and Luo, Lei},
  journal={International Journal of Computational Intelligence Systems},
  volume={17},
  number={1},
  pages={11},
  year={2024},
  publisher={Springer}
}

@article{wang2024qwen2,
  title={Qwen2-vl: Enhancing vision-language model's perception of the world at any resolution},
  author={Wang, Peng and Bai, Shuai and Tan, Sinan and Wang, Shijie and Fan, Zhihao and Bai, Jinze and Chen, Keqin and Liu, Xuejing and Wang, Jialin and Ge, Wenbin and others},
  journal={arXiv preprint arXiv:2409.12191},
  year={2024}
}

@inproceedings{long2015fully,
  title={Fully convolutional networks for semantic segmentation},
  author={Long, Jonathan and Shelhamer, Evan and Darrell, Trevor},
  booktitle={Proceedings of the IEEE conference on computer vision and pattern recognition},
  pages={3431--3440},
  year={2015}
}

@article{russakovsky2015imagenet,
  title={Imagenet large scale visual recognition challenge},
  author={Russakovsky, Olga and Deng, Jia and Su, Hao and Krause, Jonathan and Satheesh, Sanjeev and Ma, Sean and Huang, Zhiheng and Karpathy, Andrej and Khosla, Aditya and Bernstein, Michael and others},
  journal={International journal of computer vision},
  volume={115},
  number={3},
  pages={211--252},
  year={2015},
  publisher={Springer}
}

@inproceedings{lin2014microsoft,
  title={Microsoft coco: Common objects in context},
  author={Lin, Tsung-Yi and Maire, Michael and Belongie, Serge and Hays, James and Perona, Pietro and Ramanan, Deva and Doll{\'a}r, Piotr and Zitnick, C Lawrence},
  booktitle={European conference on computer vision},
  pages={740--755},
  year={2014},
  organization={Springer}
}

@article{ORENGO2026106607,
title = {High-performance {3D} morphometrics via deep learning and tabular foundation models: a case study on complex cereal grain classification},
journal = {Journal of Archaeological Science},
volume = {191},
pages = {106607},
year = {2026},
issn = {0305-4403},
author = {H.A. Orengo and I. Berganzo-Besga and J. Esmoris and F. Lumbreras and P. Aliende and M. Wallace and A. Livarda}
}

@article{Eleftheriadou-2025,
 author = {Eleftheriadou, Anastasia and McPherron, Shannon P. and Marreiros, João},
 doi = {10.5334/jcaa.190},
 journal = {Journal of Computer Applications in Archaeology},
 month = {Jun},
 title = {Machine Learning Applications in Use-Wear Analysis: A Critical Review},
 year = {2025}
}

@article{eberl2023machine,
  title={Machine learning--based identification of lithic microdebitage},
  author={Eberl, Markus and Bell, Charreau S and Spencer-Smith, Jesse and Raj, Mark and Sarubbi, Amanda and Johnson, Phyllis S and Rieth, Amy E and Chaudhry, Umang and Aguila, Rebecca Estrada and McBride, Michael},
  journal={Advances in Archaeological Practice},
  volume={11},
  number={2},
  pages={152--163},
  year={2023},
  publisher={Cambridge University Press}
}

@article{sferrazza2025archaeological,
  title={Archaeological and experimental lithic microwear classification through {2D} textural analysis and machine learning},
  author={Sferrazza, Paolo},
  journal={Journal of Archaeological Method and Theory},
  volume={32},
  number={1},
  pages={31},
  year={2025},
  publisher={Springer}
}

@incollection{ulanowskaWhyNotWool2023,
	address = {Wiesbaden},
	series = {Forschungen und {Berichte} zur {Archäologie} in {Baden}-{Württemberg}},
	title = {Why not wool? {Evidence} for raw materials and technical uses of textiles based on imprints on the undersides of clay sealings from {Bronze} {Age} {Greece}},
	number = {28},
	booktitle = {{THE} {SIGNIFICANCE} {OF} {ARCHAEOLOGICAL} {TEXTILES} {Papers} of the {Interna}-tional {Online} {Conference} 24th–25th {February} 2021. {THEFBO}, {Volume} {II}},
	publisher = {Reichert},
	author = {Ulanowska, Agata},
	editor = {Banck-Burgess, Johanna and Marinova, Elena and Mischka, Doris},
	year = {2023},
	pages = {165--177},
}

@article{10.1098/rsta.1933.0009,
    author = {Neyman, Jerzy and Pearson, Egon Sharpe},
    title = {IX. On the problem of the most efficient tests of statistical hypotheses},
    journal = {Philosophical Transactions of the Royal Society of London, Series A: Containing Papers of a Mathematical or Physical Character},
    volume = {231},
    number = {694-706},
    pages = {289-337},
    year = {1933},
    month = {02},
    issn = {0264-3952},
}

@article{cybulska,
author = {Cybulska, Maria},
year = {2007},
month = {01},
pages = {64-65},
title = {Archaeological Textiles–A Need for New Methods of Analysis and Reconstruction},
volume = {15},
journal = {Fibres and Textiles in Eastern Europe}
}

@article{outram,
  title={Introduction to experimental archaeology},
  author={Outram, Alan K},
  journal={World Archaeology},
  volume={40},
  number={1},
  pages={1--6},
  year={2008},
  publisher={Taylor \& Francis}
}

@Article{liu2022convnet,
  author  = {Zhuang Liu and Hanzi Mao and Chao-Yuan Wu and Christoph Feichtenhofer and Trevor Darrell and Saining Xie},
  title   = {A ConvNet for the 2020s},
  journal = {Proceedings of the IEEE/CVF Conference on Computer Vision and Pattern Recognition (CVPR)},
  year    = {2022},
}

@article{winterbottom2022deep,
  title={A deep learning approach to fight illicit trafficking of antiquities using artefact instance classification},
  author={Winterbottom, Thomas and Leone, Anna and Al Moubayed, Noura},
  journal={Scientific Reports},
  volume={12},
  number={1},
  pages={13468},
  year={2022},
  publisher={Nature Publishing Group UK London}
}

@inproceedings{yoshimura2019deep,
  title={Deep learning architect: classification for architectural design through the eye of artificial intelligence},
  author={Yoshimura, Yuji and Cai, Bill and Wang, Zhoutong and Ratti, Carlo},
  booktitle={International Conference on Computers in Urban Planning and Urban Management},
  pages={249--265},
  year={2019},
  organization={Springer}
}

@article{pei2024recognizing,
  title={Recognizing materials in cultural relic images using computer vision and attention mechanism},
  author={Pei, Huining and Zhang, Chuyi and Zhang, Xinxin and Liu, Xinyu and Ma, Yujie},
  journal={Expert systems with applications},
  volume={239},
  pages={122399},
  year={2024},
  publisher={Elsevier}
}

@article{elbehairy2026comprehensive,
  title={A comprehensive review of deep learning methods in damage classification, detection, and segmentation of cultural heritage sites},
  author={ElBehairy, Aya and El-Nasr, Nermean A Abu and Grimberg, Phillip and Said, Lobna A},
  journal={Journal of Cultural Heritage},
  volume={78},
  pages={228--237},
  year={2026},
  publisher={Elsevier}
}

\newpage
\appendix
\section{Supplementary Material}

\subsection{Hardware Requirements \& Technical Background} \label{app:A1}

Defining strict minimum hardware requirements is inherently difficult, as computational demands depend on user-defined configurations, including the selected computer vision models, Vision-Language Model (VLM), batch sizes, dataset sizes, and the number of concurrent users. All experiments and inference runs presented in Sec. 4, including VLM inference tasks, were executed on a workstation equipped with a single NVIDIA RTX 5090 GPU and 64~GB of system memory (RAM). The available computational resources were sufficient to execute computer vision training and inference tasks alongside the VLM pipeline for a single user. We expect single-user sequential tasks to be feasible under more constrained setups, \eg a GPU with 16~GB of VRAM and 32~GB of system memory.  While the platform itself remains hardware-agnostic and requires minimal resources for the interface and orchestration layers, several of its underlying components natively support multi-GPU execution, allowing deployments to scale across multiple GPU-enabled worker nodes. 

For deployments serving multiple users, a distributed GPU allocation strategy is recommended. In particular, dedicating a single GPU to the VLM service while scheduling computer vision training and inference workloads across separate GPU workers enables concurrent execution while preventing resource bottlenecks. As workload orchestration is handled by Kubernetes, additional GPU-enabled worker nodes can be provisioned to accommodate increasing computational demands and larger numbers of simultaneous users. The same hardware-agnostic scaling is applied to storage too: training and inference are executed on the local node, therefore the maximum practical dataset size is dictated by that node's storage capabilities rather than by AmalthAI itself.

While using the platform requires no technical background, deployment and setup assume mild familiarity with the command line and containerization. The provided documentation aims to offer a streamlined and easy process for basic setup, while deeper configuration, such as changing the available ML models or a custom data source configuration, requires the corresponding domain expertise.

\subsection{Visual Language Model Configuration and Outputs} \label{app:A3}

\begin{figure}[t]
    \centering
    \captionsetup[subfigure]{font=footnotesize, skip=1pt}

    \begin{minipage}[t]{0.56\linewidth}
        \centering
        \includegraphics[width=\linewidth]{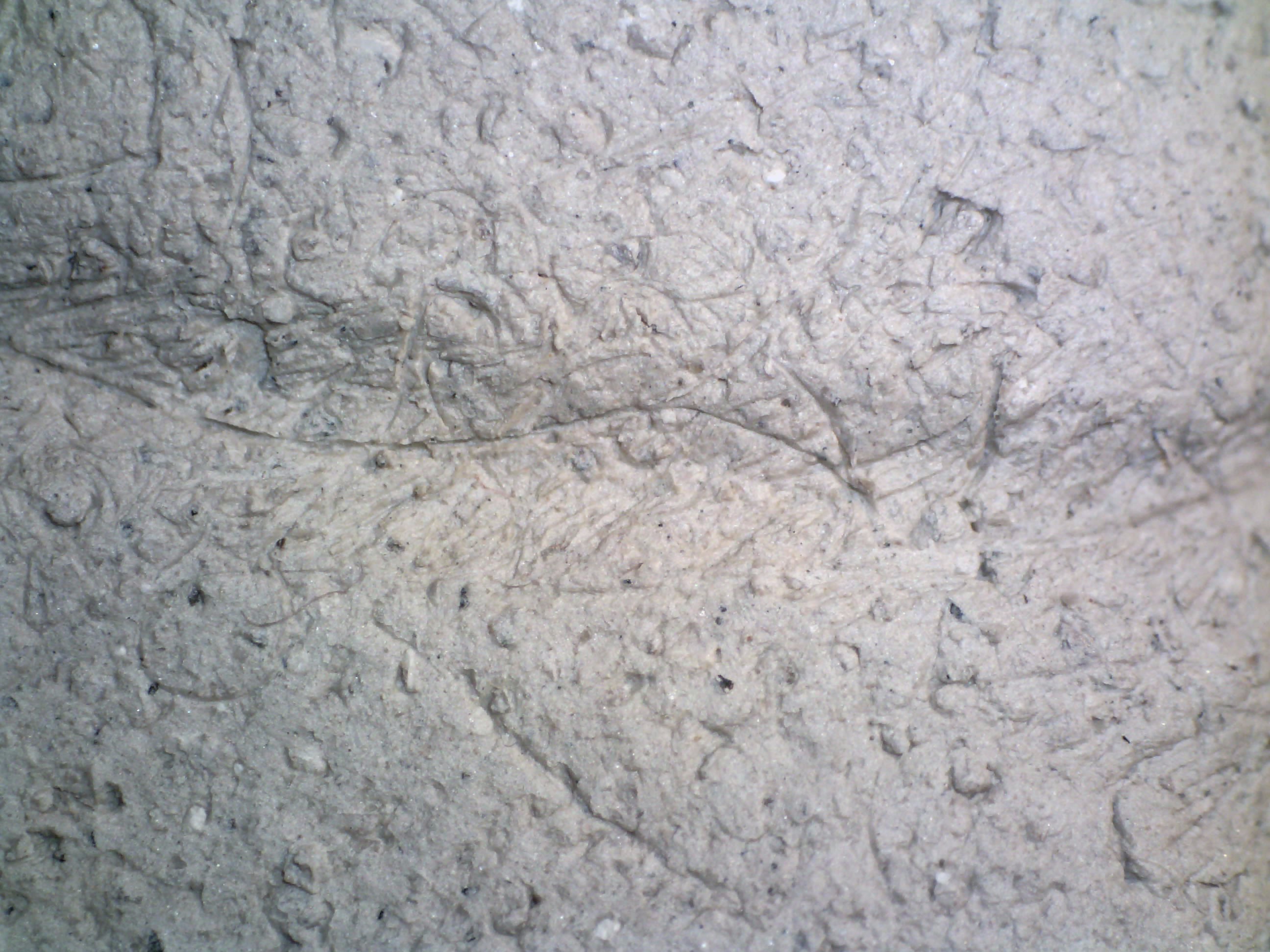}
        \subcaption{Raw microscope image}
    \end{minipage}
    \hfill
    \begin{minipage}[t]{0.42\linewidth}
        \centering
        \includegraphics[width=\linewidth]{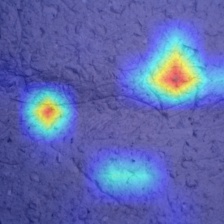}
        \subcaption{Grad-CAM}
    \end{minipage}

    \vspace{0.25em}

    \begin{promptbox}
Input prompt:
Task: material classification

The attached images are:
1. The original Cultural Heritage image.
2. The corresponding Grad-CAM visualization.

Known classification results (do NOT predict or re-derive them):
  Ground-truth class: Wool
  Predicted class: Nettle
  Prediction confidence: 0.4438
  Class probabilities:
    Flax: 0.37%
    Nettle: 44.38%
    Tree bast_lime: 34.35%
    Wool: 20.90%

Using the original image and the Grad-CAM, explain why the classifier may have predicted the predicted class instead of the ground-truth class.
Base your explanation only on visible evidence in the images.
Use the class probabilities only as supporting information.
Return only the JSON specified in the system prompt.

Output:
{
  "visible_description": "The image shows a textured surface with a central area that appears to be a textile imprint. The surface is light grey and has a rough texture with some darker spots and lines.",
  "attention_description": "The heat is concentrated in the central area of the image, which is the textile imprint. The heat is brighter and more intense in the central area compared to the surrounding areas, indicating that the classifier focused on this area.",
  "task_properties": {
    "task": "material classification",
    "ground_truth": "Wool",
    "predicted_class": "Nettle",
    "confidence": 0.4438,
    "class_probabilities": {
      "Flax": 0.37,
      "Nettle": 44.38,
      "Tree bast lime": 34.35,
      "Wool": 20.90
    }
  }
}
    \end{promptbox}

    \caption{Example of a visual explanation pipeline for material-related errors. The model processes a raw microscope image (a), its Grad-CAM heatmap (b), and prediction metadata to produce a structured JSON explanation of a misclassified material.}
    \label{fig:vlm_output}
\end{figure}

\begin{figure}[t]
    \centering
    \captionsetup[subfigure]{font=footnotesize, skip=1pt}

    \begin{minipage}[t]{0.56\linewidth}
        \centering
        \includegraphics[width=\linewidth]{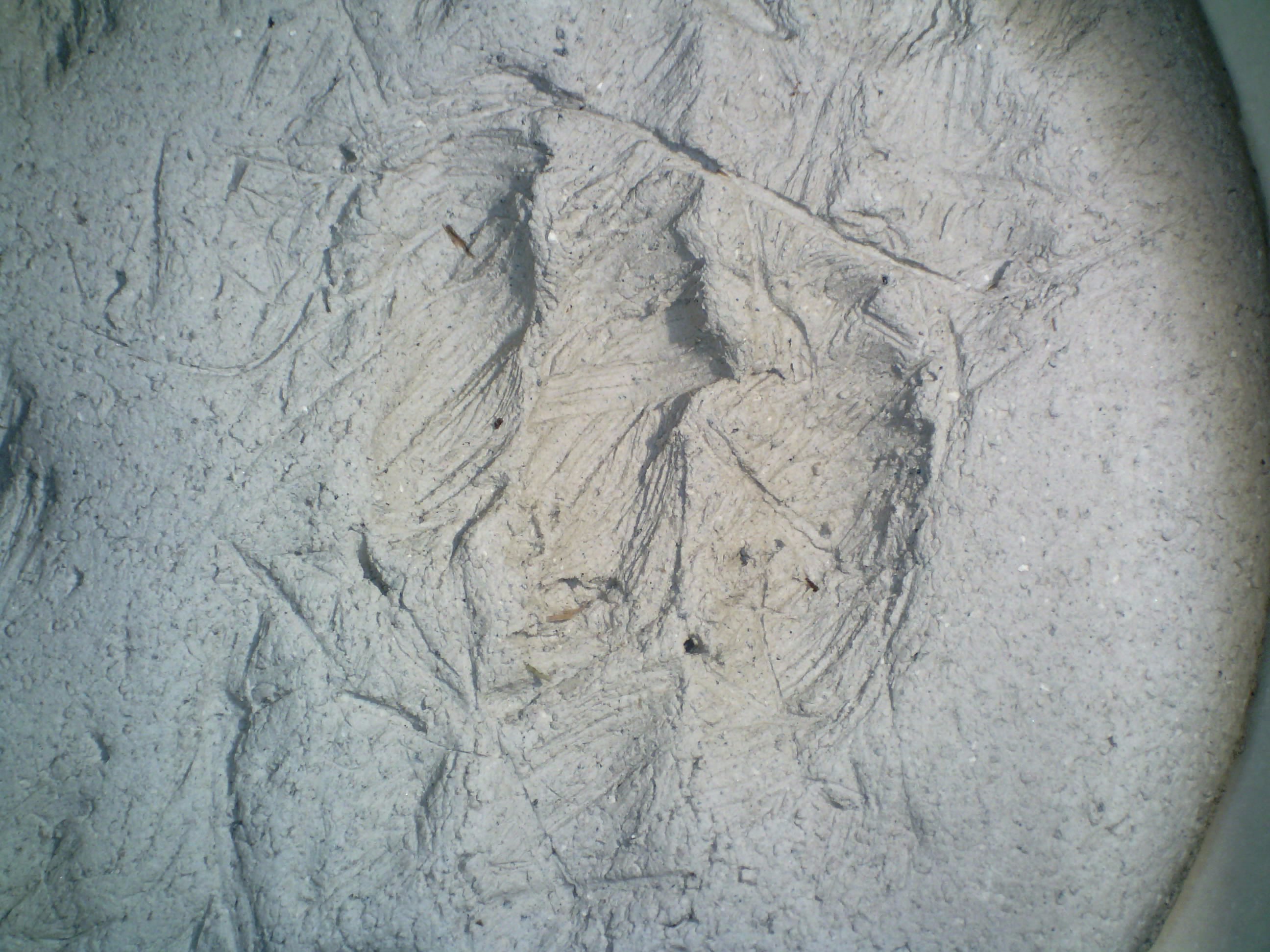}
        \subcaption{Raw microscope image}
    \end{minipage}
    \hfill
    \begin{minipage}[t]{0.42\linewidth}
        \centering
        \includegraphics[width=\linewidth]{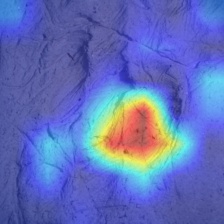}
        \subcaption{Grad-CAM}
    \end{minipage}

    \vspace{0.25em}

    \begin{promptbox}
Input prompt:
Task: material classification

The attached images are:
1. The original Cultural Heritage image.
2. The corresponding Grad-CAM visualization.

Known classification results (do NOT predict or re-derive them):
  Ground-truth class: Nettle
  Predicted class: Wool
  Prediction confidence: 0.6959
  Class probabilities:
    Flax: 0.25%
    Nettle: 20.73%
    Tree bast_lime: 9.43%
    Wool: 69.59%

Using the original image and the Grad-CAM, explain why the classifier may have predicted the predicted class instead of the ground-truth class.
Base your explanation only on visible evidence in the images.
Use the class probabilities only as supporting information.
Return only the JSON specified in the system prompt.

Output:
{
  "visible_description": "The image shows a clay imprint made by pressing textile samples into clay under controlled conditions. The imprint is circular and contains visible threads and fibers, indicating the presence of textile material. The background is a light-colored, textured surface, likely clay or a similar material.",
  "attention_description": "The heat in the Grad-CAM visualization is concentrated in the center of the image, near the circular imprint. This suggests that the classifier focused on the central area of the imprint, which is likely the textile material. The surrounding area, including the background, does not show any significant heat, indicating that the classifier did not focus on these areas.",
  "task_properties": {
    "task": "material classification",
    "ground_truth": "Nettle",
    "predicted_class": "Wool",
    "confidence": 0.6959,
    "class_probabilities": {
      "Flax": 0.25,
      "Nettle": 20.73,
      "Tree bast lime": 9.43,
      "Wool": 69.59
    }
  }
}
    \end{promptbox}

    \caption{Example of a visual explanation pipeline for material-related errors. The model processes a raw microscope image (a), its Grad-CAM heatmap (b), and prediction metadata to produce a structured JSON explanation of a misclassified material.}
    \label{fig:vlm_output1}
\end{figure}

\begin{figure}[t]
    \centering
    \captionsetup[subfigure]{font=footnotesize, skip=1pt}

    \begin{minipage}[t]{0.62\linewidth}
        \centering
        \includegraphics[width=\linewidth]{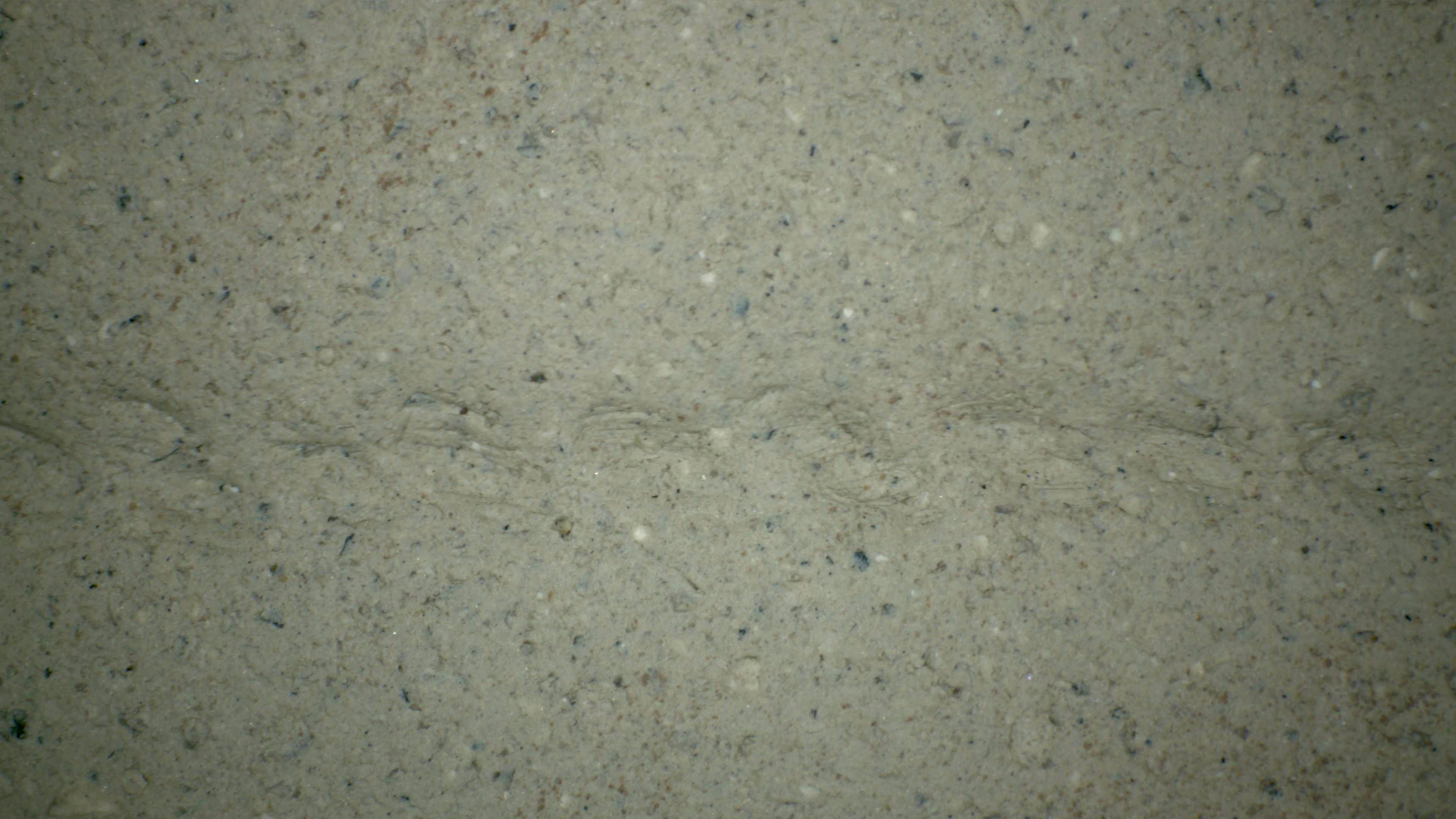}
        \subcaption{Raw microscope image}
    \end{minipage}
    \hfill
    \begin{minipage}[t]{0.35\linewidth}
        \centering
        \includegraphics[width=\linewidth]{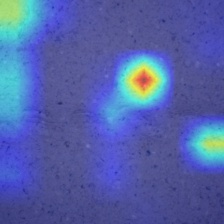}
        \subcaption{Grad-CAM}
    \end{minipage}

    \vspace{0.25em}

    \begin{promptbox}
Input prompt:
Task: technique classification

The attached images are:
1. The original Cultural Heritage image.
2. The corresponding Grad-CAM visualization.

Known classification results (do NOT predict or re-derive them):
  Ground-truth class: Drilling
  Predicted class: Spinning
  Prediction confidence: 0.5679
  Class probabilities:
    Drilling: 37.99%
    Spinning: 56.79%
    Splicing: 5.21%

Using the original image and the Grad-CAM, explain why the classifier may have predicted the predicted class instead of the ground-truth class.
Base your explanation only on visible evidence in the images.
Use the class probabilities only as supporting information.
Return only the JSON specified in the system prompt.

Output:
{
  "visible_description": "The image shows a textured surface with a central focus area that appears to be a small, circular object with a bright spot in the center.",
  "attention_description": "The bright spot in the center of the image is spatially concentrated in the area of the textile imprint, indicating that the classifier may have focused on this area due to its high probability of being a specific feature or object.",
  "task_properties": {
    "task": "technique classification",
    "ground_truth": "Drilling",
    "predicted_class": "Spinning",
    "confidence": 0.5679,
    "class_probabilities": {
      "Drilling": 37.99,
      "Spinning": 56.79,
      "Splicing": 5.21
    }
  }
}
    \end{promptbox}

    \caption{Example of a visual explanation pipeline for errors in technique recognition. A raw microscope image (a), its corresponding Grad-CAM visualization (b), and classification outputs are combined to generate a structured JSON explanation of a technique-level misclassification.}
    \label{fig:vlm_output2}
\end{figure}

\begin{figure}[t]
    \centering
    \captionsetup[subfigure]{font=footnotesize, skip=1pt}

    \begin{minipage}[t]{0.62\linewidth}
        \centering
        \includegraphics[width=\linewidth]{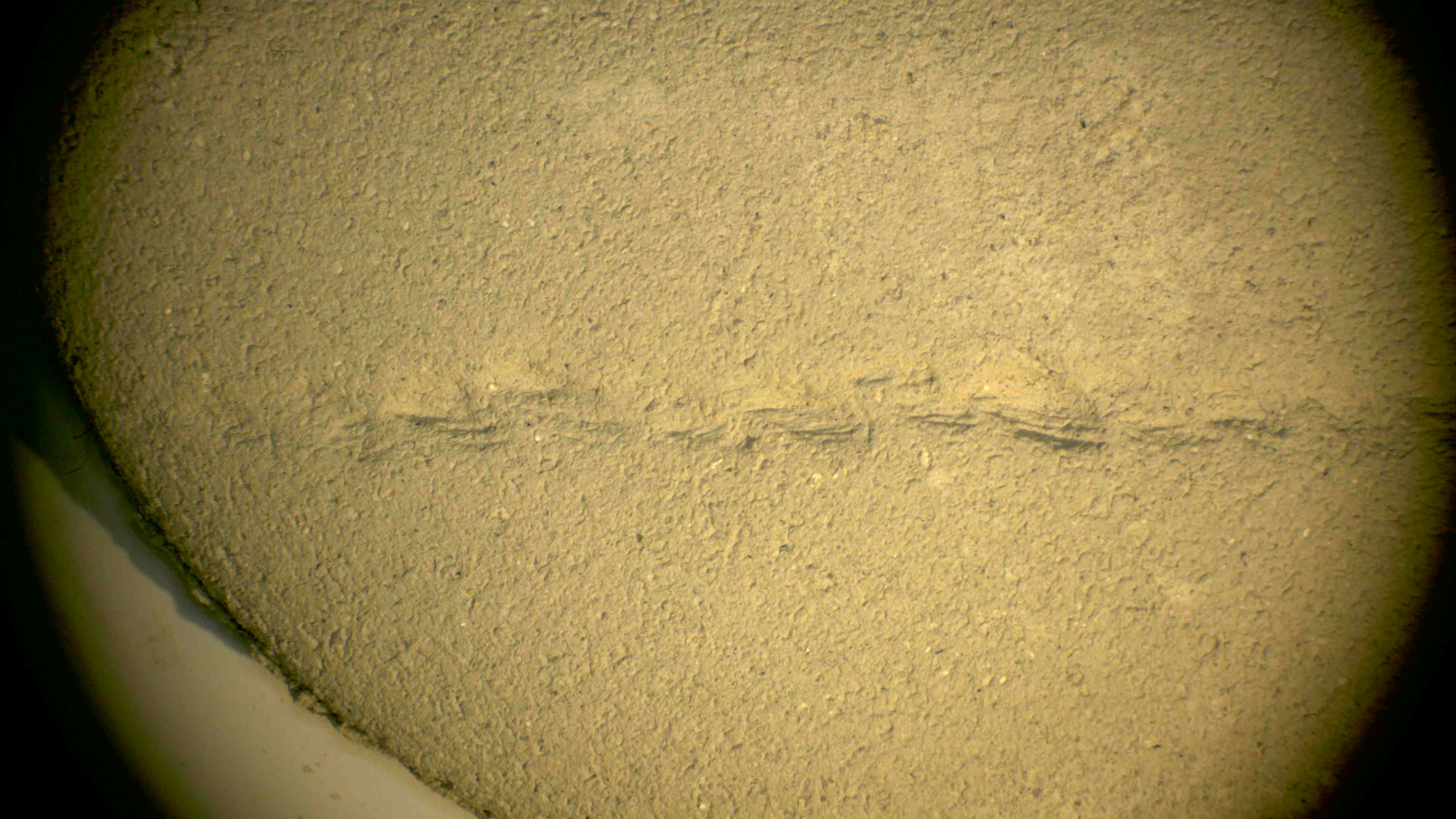}
        \subcaption{Raw microscope image}
    \end{minipage}
    \hfill
    \begin{minipage}[t]{0.35\linewidth}
        \centering
        \includegraphics[width=\linewidth]{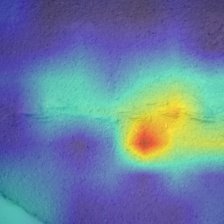}
        \subcaption{Grad-CAM}
    \end{minipage}

    \vspace{0.25em}

    \begin{promptbox}
Input prompt:
Task: technique classification

The attached images are:
1. The original Cultural Heritage image.
2. The corresponding Grad-CAM visualization.

Known classification results (do NOT predict or re-derive them):
  Ground-truth class: Drilling
  Predicted class: Splicing
  Prediction confidence: 0.3488
  Class probabilities:
    Drilling: 31.39%
    Spinning: 33.73%
    Splicing: 34.88%

Using the original image and the Grad-CAM, explain why the classifier may have predicted the predicted class instead of the ground-truth class.
Base your explanation only on visible evidence in the images.
Use the class probabilities only as supporting information.
Return only the JSON specified in the system prompt.

Output:
{
  "visible_description": "The image shows a textured surface with visible imprints of textile samples pressed into clay.",
  "attention_description": "The heat is concentrated in the center of the image, indicating the area where the textile imprint is located.",
  "task_properties": {
    "task": "technique classification",
    "ground_truth": "Drilling",
    "predicted_class": "Splicing",
    "confidence": 0.3488,
    "class_probabilities": {
      "Drilling": 31.39,
      "Spinning": 33.73,
      "Splicing": 34.88
    }
  }
}
    \end{promptbox}

    \caption{Example of a visual explanation pipeline for errors in technique recognition. A raw microscope image (a), its corresponding Grad-CAM visualization (b), and classification outputs are combined to generate a structured JSON explanation of a technique-level misclassification.}
    \label{fig:vlm_output3}
\end{figure}

Beyond the three core computer vision tasks, AmalthAI provides a Vision-Language Model (VLM) tool for interpreting classification results. For the use case described above, users can inspect individual misclassified samples by prompting the VLM with the original raw microscope image, the corresponding Grad-CAM visualization, which is generated from lower-resolution feature maps during inference, and the classifier's prediction metadata, including the ground-truth label, predicted label, prediction confidence, and class probability distribution. The VLM is guided by a dedicated system prompt (Fig.~\ref{fig:gradcam_system_prompt}), which instructs it to explain why the classifier may have favored the predicted class over the ground-truth class based solely on the provided visual evidence, without re-evaluating the prediction. To encourage consistent reasoning and a standardized JSON response, the prompt is further augmented with four representative in-context examples extracted from the classification inference stage (Figs.~\ref{fig:vlm_output}--\ref{fig:vlm_output3}).

\begin{figure}[ht]
    \centering
    \begin{minipage}{1\linewidth}
    \begin{systempromptbox}
    \begin{lstlisting}[style=systempromptstyle,basicstyle=\ttfamily\tiny,breaklines=true]
SYSTEM_PROMPT = """
You are a visual analysis assistant helping cultural heritage experts understand microscope images from an experimental archaeology study.
The images show clay imprints made by pressing textile samples (fibres and threads) into clay under controlled conditions.

You will receive TWO images:
1. Image 1 (raw): the actual microscope photograph of the clay imprint.
2. Image 2 (Grad-CAM): a heatmap overlay showing where the classifier focused.

"Rules:\n"
    "- visible_description: describe ONLY what you see in Image 1. "
    "Do NOT mention the heatmap or Grad-CAM here.\n"
    "- attention_description: describe WHERE the heat in Image 2 is spatially "
    "(center / edge / background / specific feature). "
    "State whether it overlaps the textile imprint. Do NOT mention colors.\n"
    "- Do NOT name any class, technique, or material type (no Drilling, Spinning, Nettle, Wool, etc.)\n"
    "- Do NOT say the classifier was right or wrong.\n"
    "- Do NOT describe the material or the background as concrete, stone, or fossil.\n"
    "- Do NOT use information about classes, techniques, or materials to inform your descriptions.\n"
    "Return ONLY valid JSON, 1-2 sentences per field:\n"
    '{
  "visible_description": "",
  "attention_description": "",
  "task_properties": {
    "task": "",
    "ground_truth": "",
    "predicted_class": "",
    "confidence": "",
    "class_probabilities": {}
  }
}'
"""
    \end{lstlisting}
    \end{systempromptbox}
    \end{minipage}%
    \caption{System prompt for controlled visual explanations, separating raw-image description from Grad-CAM attention while preventing class labels, correctness judgments, and unsupported interpretations.}
    \label{fig:gradcam_system_prompt}
\end{figure}

\FloatBarrier
\subsection{Platform User Interface Visualizations} \label{app:A4}

\begin{figure}[!htbp]
    \centering

    \begin{subfigure}[t]{0.60\linewidth}
        \centering
        \includegraphics[width=\linewidth]{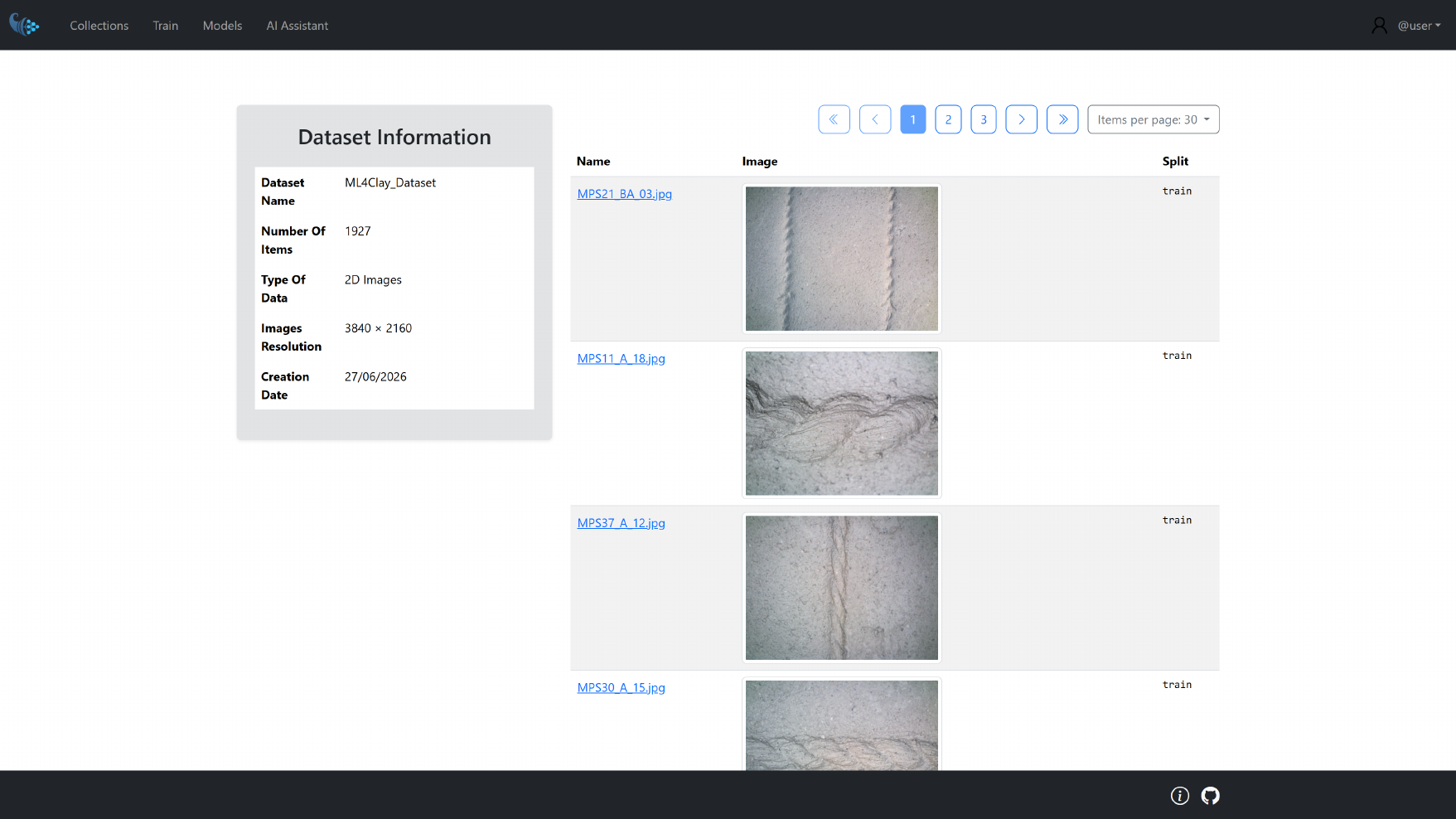}
        \caption{Dataset information interface.}
        \label{fig:ui_dataset}
    \end{subfigure}

    \vspace{0.75em}

    \begin{subfigure}[t]{0.60\linewidth}
        \centering
        \includegraphics[width=\linewidth]{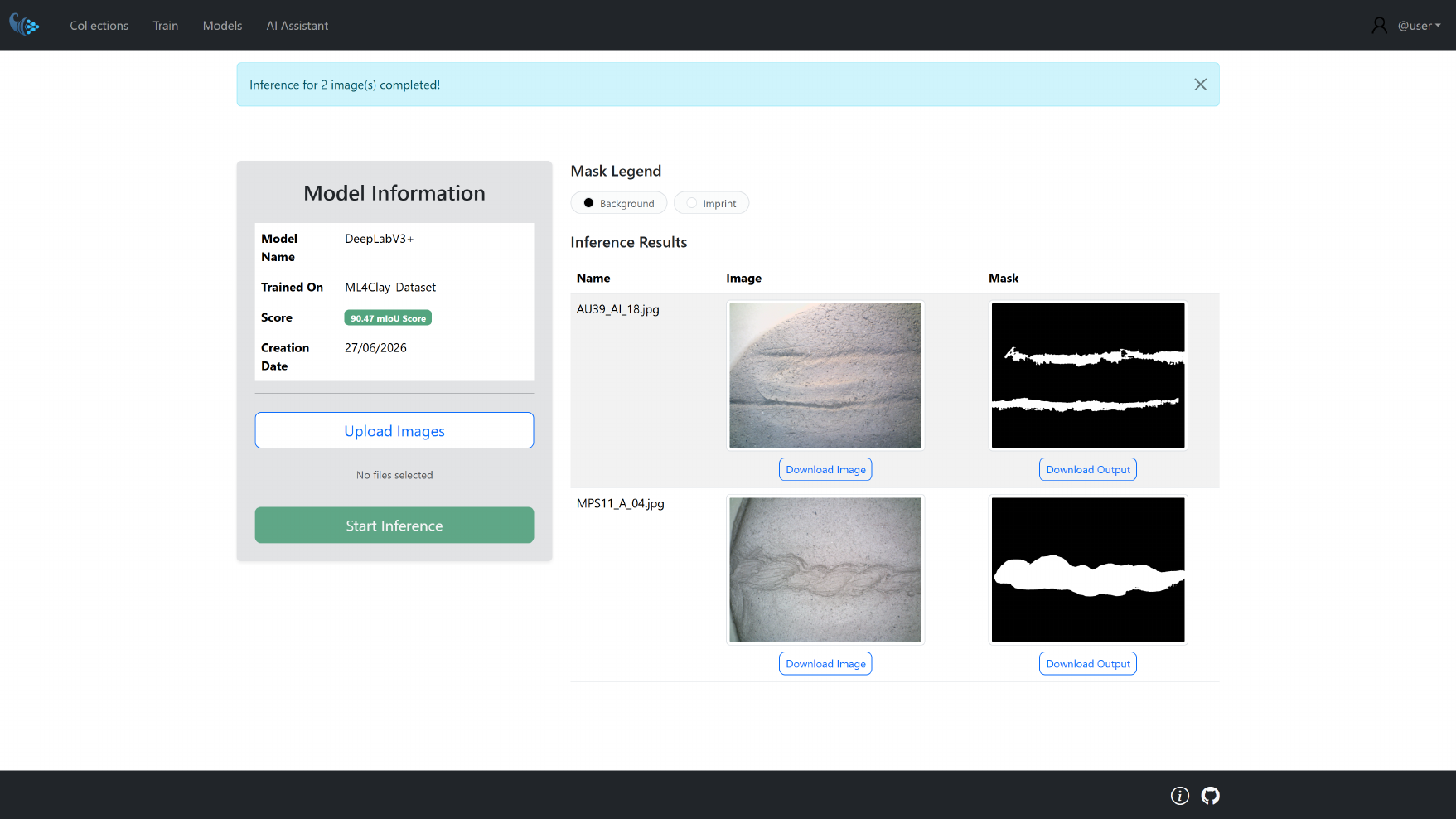}
        \caption{Inference interface.}
        \label{fig:ui_inference}
    \end{subfigure}

    \vspace{0.75em}

    \begin{subfigure}[t]{0.60\linewidth}
        \centering
        \includegraphics[width=\linewidth]{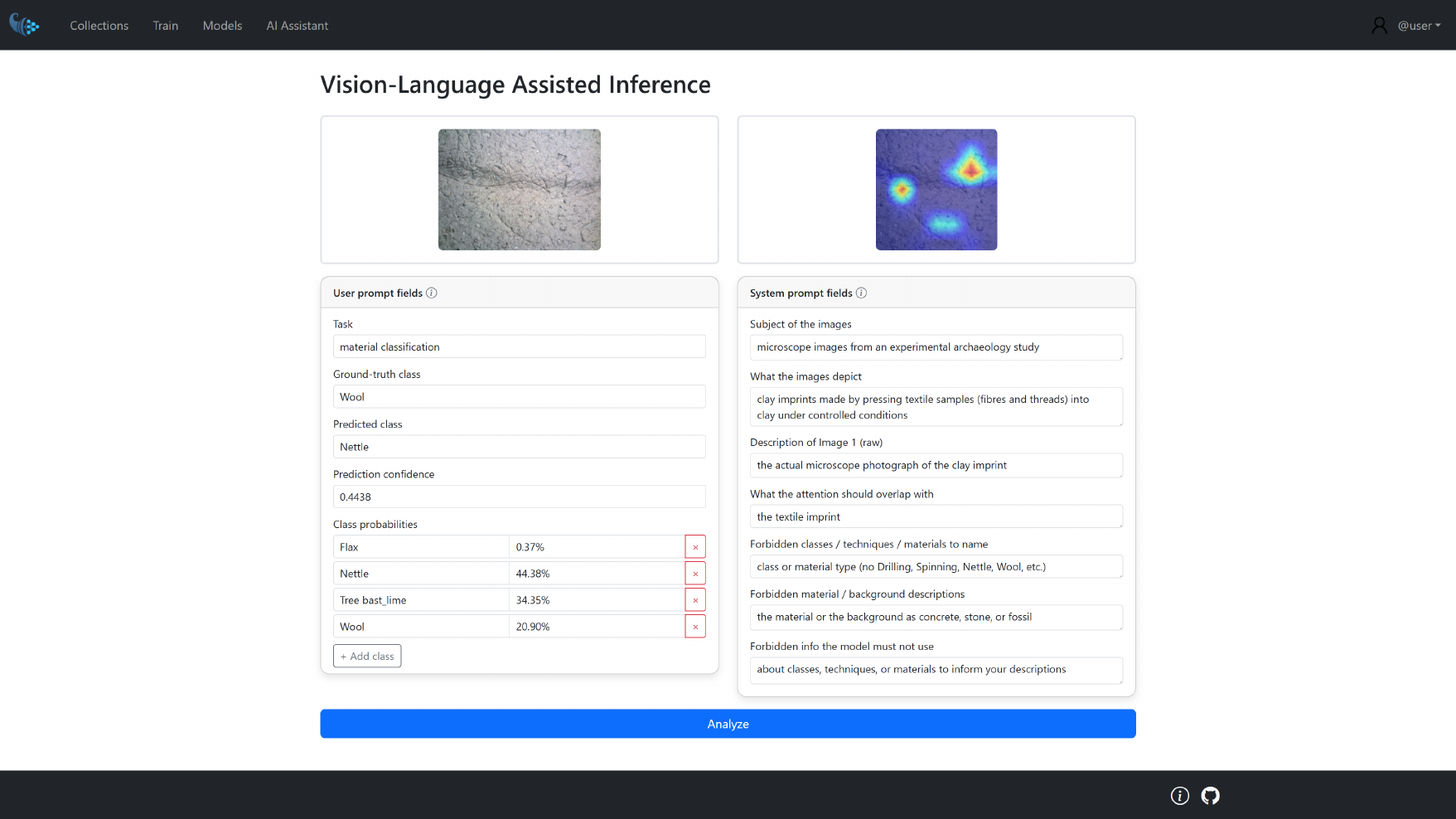}
        \caption{VLM interface.}
        \label{fig:ui_vlm}
    \end{subfigure}

    \caption{Platform user-interface visualizations in AmalthAI. The panels show the dataset information interface, the inference interface, and the AI assistant interface.}
    \label{fig:amalthai_ui}
\end{figure}

Fig.~\ref{fig:amalthai_ui} summarizes the main AmalthAI user-interface visualizations. The dataset interface supports inspection of uploaded data splits and dataset attributes, the inference interface presents model outputs and segmentation masks, and the AI Assistant page provides interaction with a VLM that supports visual-language inspection of classification results.

\FloatBarrier

\subsection{Qualitative Segmentation Results} \label{app:A2}

\begin{figure}[ht]
    \centering
    \includegraphics[width=0.99\linewidth]{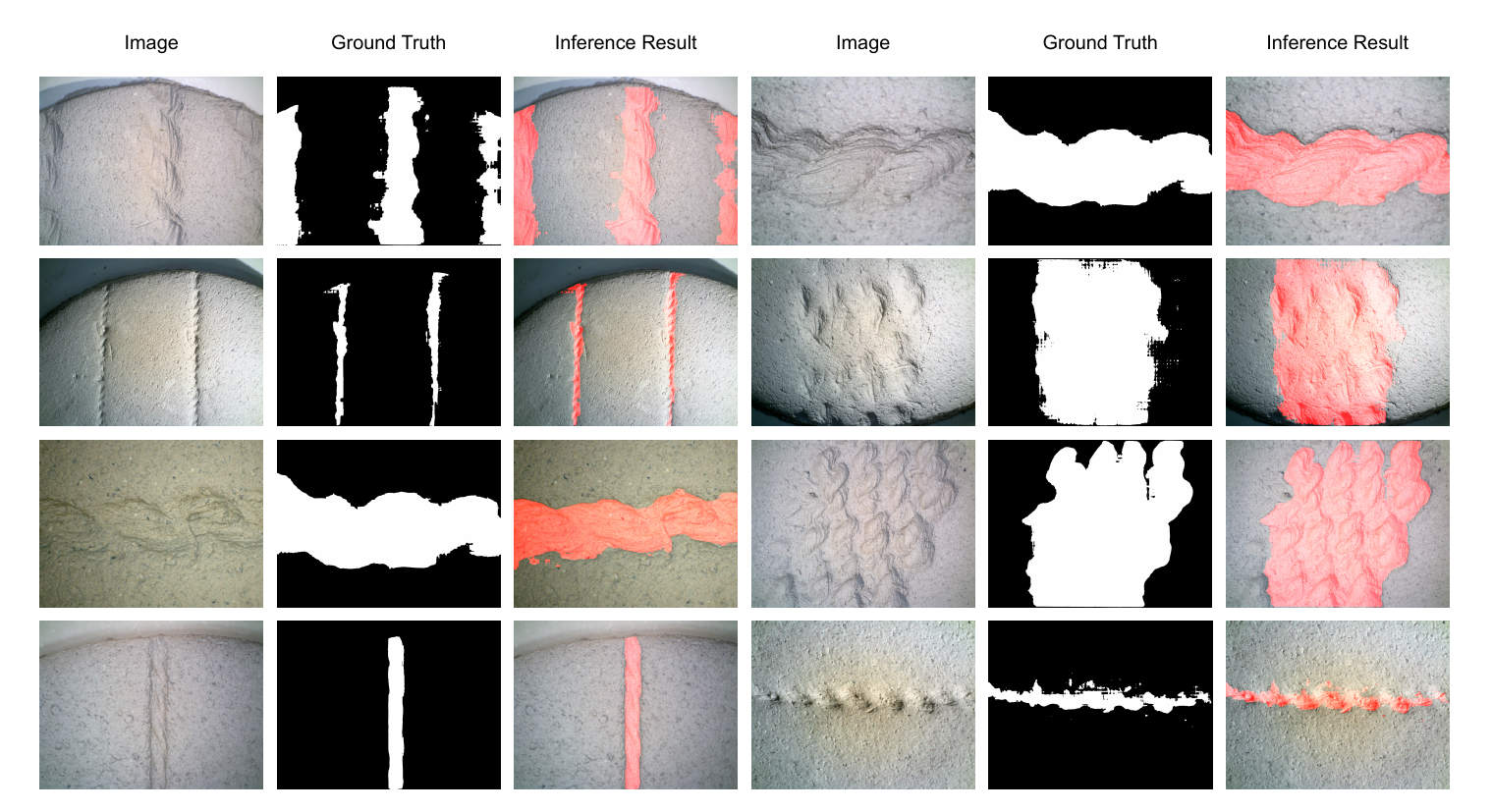}
    \caption{Qualitative segmentation inference results. The inference was performed by the best performing model across the experimental textile imprint dataset. Predicted segmentation masks are colored in red.}
    \label{fig:qualitative_segmentation}
\end{figure}

Fig.~\ref{fig:qualitative_segmentation} presents qualitative segmentation results for the textile-imprint dataset. These examples complement the quantitative segmentation results by illustrating how the learned masks can support downstream expert inspection of the relevant imprint area.

\end{document}